\documentclass{article}

\usepackage[preprint]{neurips_2025}

\usepackage[utf8]{inputenc} 
\usepackage[T1]{fontenc}    
\usepackage{hyperref}       
\usepackage{url}            
\usepackage{graphicx}       
\usepackage{booktabs}       
\usepackage{colortbl}       
\usepackage{array}          
\usepackage{capt-of}        
\usepackage{nicefrac}       
\usepackage{microtype}      
\usepackage{xcolor}         
\usepackage{xspace}
\usepackage{amsmath,amsfonts}

\title{Self-Evolving Embodied Agents via Skill-Harness Evolution}

\newcommand{\equalmark}{\ensuremath{*}}
\newcommand{\corrmarkA}{\ensuremath{\dagger}}
\newcommand{\corrmarkB}{\ensuremath{\ddagger}}
\newcommand{\internmark}{\S}

\author{%
  \textbf{Peidong Wang}\textsuperscript{1,2,\equalmark,\internmark}
  \quad
  \textbf{Zhiming Ma}\textsuperscript{1,\equalmark}
  \quad
  \textbf{Ying Chang}\textsuperscript{2,\equalmark,\internmark}
  \quad
  \textbf{Xufang Luo}\textsuperscript{2,\corrmarkA}
  \quad
  \textbf{Yiqun Zhang}\textsuperscript{1}
  \\
  \textbf{Zihan Wang}\textsuperscript{1}
  \quad
  \textbf{Xiaocui Yang}\textsuperscript{1}
  \quad
  \textbf{Shi Feng}\textsuperscript{1,\corrmarkB}
  \quad
  \textbf{Yuqing Yang}\textsuperscript{2}
  \quad
  \textbf{Dongsheng Li}\textsuperscript{2}
  \\[0.5em]
  \normalfont\small\textsuperscript{1}Northeastern University
  \\
  \normalfont\small\textsuperscript{2}Microsoft Research
}

\newcommand{\ourmethod}{SHAPER\xspace}

\begin{document}

\maketitle
\begingroup
\renewcommand{\thefootnote}{\fnsymbol{footnote}}
\footnotetext[1]{Equal contribution.}
\footnotetext[2]{Corresponding author: \texttt{luoxufang@outlook.com}.}
\footnotetext[3]{Corresponding author: \texttt{fengshi@cse.neu.edu.cn}.}
\footnotetext[4]{Work done while Peidong Wang and Ying Chang were interns at
Microsoft Research Asia.}
\endgroup

\begin{abstract}
  Embodied agents are increasingly built as systems around foundation models, where performance depends not only on model weights but also on the skills, context, action interfaces, and execution harness surrounding the model.
While supervised fine-tuning and reinforcement learning can adapt agents to new environments, they require additional data, rewards, and training runs; meanwhile, many train-free code-centric approaches rely on programmable robot APIs that may be unavailable in fixed-interface settings.
We propose \ourmethod, a self-evolving framework for train-free embodied adaptation that keeps model parameters frozen and improves the non-parametric agent system by evolving reusable skills and a context-code harness through target-environment rollouts.
In \ourmethod, the same frozen model can serve as both planner and optimizer, refining its external skills and context code harness without parameter updates.
We evaluate \ourmethod on VLABench and ESI-Bench, covering embodied agents with different low-level action interfaces, and compare against pure execution, supervised fine-tuning, and test-time-scaling baselines such as verifier-free
selection and voting.
Our results suggest that skill-and-harness optimization is a practical route to self-evolving embodied agents when model training is expensive, unavailable, or undesirable.

\end{abstract}

\section{Introduction}

Embodied AI is increasingly moving from isolated policy models toward agent systems.
Recent vision-language-action models and multimodal foundation models have shown that large pretrained models can perceive scenes, interpret instructions, reason over objects and spatial relations, and act in simulated or physical environments~\cite{kim2024openvla,black2024pi0,black2025pi05,physicalintelligence2026pi07,bjorck2025grootn1,gemini2025robotics,yuan2026qwenrobotmanip}.
At the same time, work on agent scaffolds, agent-computer interfaces, harness optimization, and robot skill discovery shows that an agent's behavior is not determined by the model alone~\cite{yang2024sweagent,chen2026agentspec,pan2026naturalharness,lee2026metaharness,lou2026autoharness,ju2026embodiskill,lu2026aspire}.
A deployed embodied agent is shaped by model-external components, including reusable skills, context construction, action interfaces, output parsing, and execution wrappers.
These non-parametric choices are especially consequential in embodied settings, where an agent must act through constrained action interfaces under partial visual feedback.

\begin{figure}[t]
    \centering
    \includegraphics[width=0.85\columnwidth]{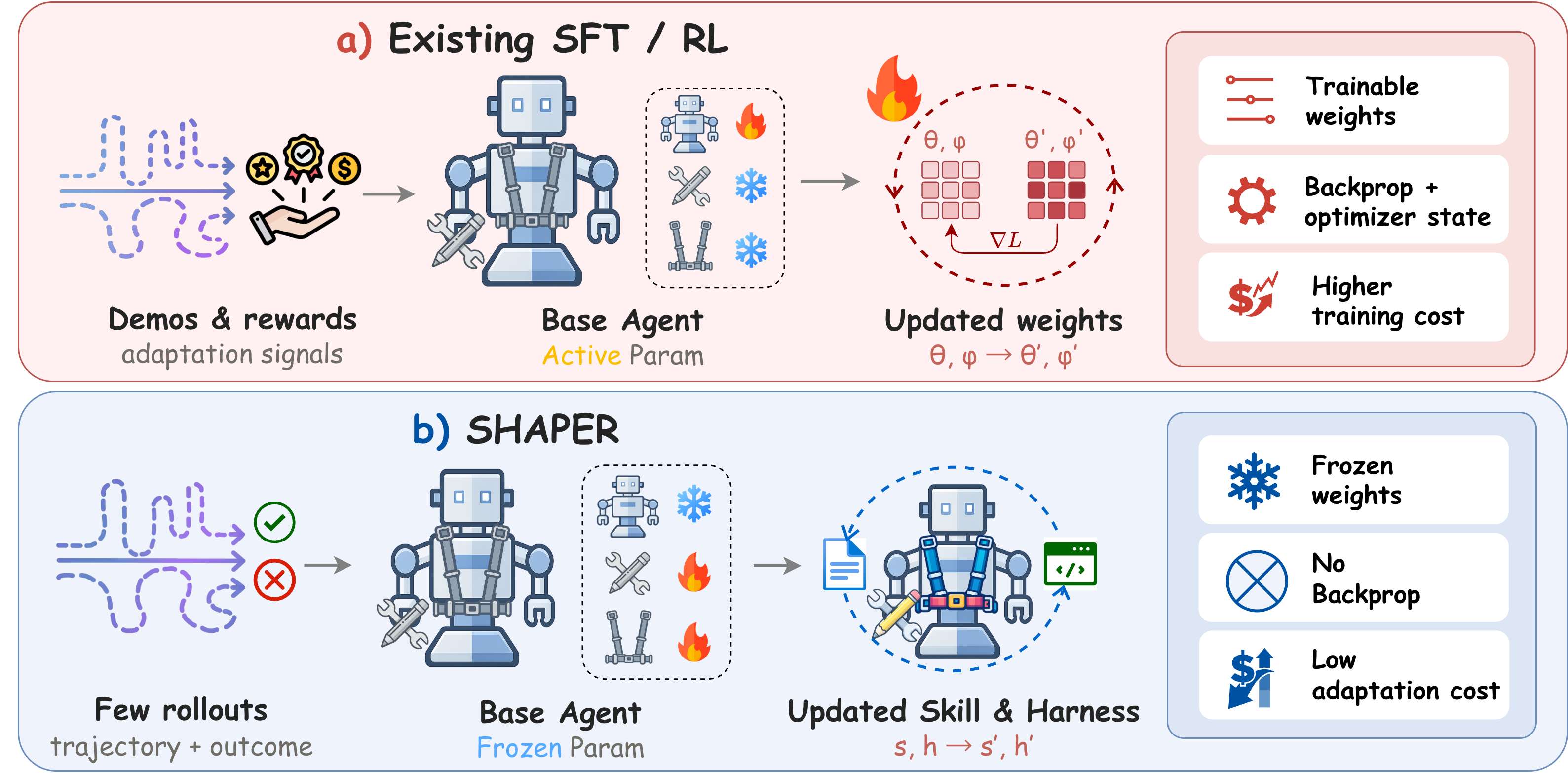}
    \caption{\ourmethod adapts frozen embodied agents by evolving skills and harnesses from a few rollouts.}
    \label{fig:teaser}
\end{figure}

A common approach to adapt embodied agents is still to update model parameters.
Supervised Fine-Tuning (SFT) and Reinforcement-Learning (RL) post-training can improve embodied performance~\cite{openx2023,khazatsky2024droid,guo2025onlinevla,lu2025vlarl,pan2026sop,physicalintelligence2025pistar06}, but require access to model weights, task-specific demonstrations or rewards, and additional optimization. These requirements limit adaptation when model weights are unavailable or interaction data are scarce. This motivates a complementary question: how can a strong frozen embodied agent be adapted to a target environment without parameter updates?

Train-free embodied adaptation offers a different path.
Recent agentic robotics systems show that procedural knowledge can live outside the model: language models can write robot programs, revise failed executions, and accumulate reusable skills without directly changing neural weights~\cite{ahn2022saycan,liang2022code,huang2023voxposer,wang2023voyager,ju2026embodiskill,lu2026aspire}.
However, much of this progress treats executable code as the action substrate.
The agent improves by generating or repairing programs that call robot-specific APIs~\cite{liang2022code,lu2026aspire}, which is powerful when such APIs, debugging hooks, and execution monitors are available.
In many embodied environments, these interfaces may be absent, incomplete, or difficult to expose, and the agent may instead have to act through a fixed or restricted action space.
This leaves open how to systematically adapt embodied agents by improving the planner-side procedural guidance and context that shape behavior, rather than by synthesizing low-level control programs.

As illustrated in Figure~\ref{fig:teaser}, we propose \ourmethod, a self-evolving framework for train-free embodied adaptation.
The key distinction is what adaptation changes.
SFT and RL turn demonstrations or rewards into updates to model weights through backpropagation; \ourmethod instead turns a small set of target-environment trajectories and outcomes into revisions of reusable \emph{skills} and a context-code \emph{harness}, while keeping the planner and executor weights frozen.
Here, a skill is a piece of procedural guidance that may be written as natural language, an action recipe, a failure-recovery rule, or a task-decomposition strategy.
The context-code harness determines which observations, prior actions, execution outcomes, and feedback are presented to the planner and how they are organized.
We reuse the same frozen VLM checkpoint in two distinct roles. During environment interaction, it acts as the upper-level planner. During evolution, it is prompted separately as an artifact optimizer that revises the textual skill and context-code harness from summarized rollout feedback. The two roles share model weights but use different prompts and inputs.
Thus, the agent evolves its external skills and context construction while keeping its model parameters fixed.

We evaluate \ourmethod across embodied environments with different low-level action interfaces.
In VLABench~\cite{zhang2024vlabench}, the upper-level VLM agent is connected to the environment through a VLA actor that serves as a low-level execution tool.
In ESI-Bench~\cite{hong2026esibench}, the same agentic formulation acts through the benchmark's environment action interface.
Across these settings, we compare against pure execution, same-data SFT, and Test-Time-Scaling (TTS) baselines such as verifier-free selection and trajectory voting~\cite{jang2025mgselect,lin2025vote}.
Our results aim to show that evolving skills and harnesses is a practical way to improve embodied agents when parameter updates are expensive, unavailable, or undesirable.

This paper makes three contributions:
(1) we formulate train-free embodied adaptation as non-parametric optimization of an embodied agent system, where reusable skills and context-code harnesses are optimized around a frozen model;
(2) we introduce \ourmethod, a self-evolving framework that improves an embodied agent by evolving textual skills and a context-code harness through target-environment rollouts, without updating model parameters;
and (3) we evaluate \ourmethod across embodied environments with different action interfaces, showing that skill-and-harness optimization can improve embodied performance and provide a competitive alternative to fine-tuning and sampling-heavy baselines.

\section{Related Work}

\paragraph{Embodied foundation models and parameter-updating adaptation.}
Recent embodied foundation models connect large-scale visual and language pretraining to robot control.
RT-2, OpenVLA, Octo, and $\pi_0$ establish generalist VLA policies, while $\pi_{0.5}$, $\pi_{0.7}$, GR00T N1, Gemini Robotics, and Qwen-RobotManip extend open-world generalization, steerability, humanoid control, and embodied reasoning~\cite{brohan2023rt2,kim2024openvla,octo2024octo,black2024pi0,black2025pi05,physicalintelligence2026pi07,bjorck2025grootn1,gemini2025robotics,yuan2026qwenrobotmanip}.
Large robot datasets and parameter-updating adaptation through supervised or reinforcement learning further broaden their capabilities~\cite{openx2023,khazatsky2024droid,guo2025onlinevla,lu2025vlarl,pan2026sop,shi2026beyondimitation,physicalintelligence2025pistar06}.
These approaches remain dependent on additional data, rewards or corrections, and model updates; we instead adapt frozen agents through external skill and harness artifacts.

\paragraph{Self-evolving agents.}
Recent work adapts frozen agents by editing model-external artifacts.
SkillOpt optimizes a persistent natural-language skill through bounded edits and
validation-gated rollouts while keeping the execution harness fixed~\cite{yang2026skillopt}.
EmbodiSkill brings skill evolution to embodied environments by distinguishing
defective skill content from execution lapses, but likewise updates the skill
rather than the context harness~\cite{ju2026embodiskill}.
AutoHarness instead synthesizes a code harness, or even a complete code policy,
from environment feedback in text-game environments, without co-optimizing
reusable procedural guidance~\cite{lou2026autoharness}.
AgentSpec represents embodied scaffolds as typed, swappable compositions for
controlled analysis rather than rollout-driven artifact optimization~\cite{chen2026agentspec}.
In contrast, \ourmethod evolves both the reusable skill and context-code harness
of an embodied agent from environment interaction, while keeping all model
parameters frozen.

\paragraph{Train-free adaptation for embodied agents.}
Train-free embodied systems compose frozen models with affordances, executable robot APIs, spatial value maps, or reusable skills~\cite{ahn2022saycan,liang2022code,huang2023voxposer,ju2026embodiskill,lu2026aspire}.
Code- and API-centric approaches, however, depend on the programmable interfaces, perception modules, and execution monitors available on each platform.
For VLA policies, vision-language steering, latent prompt optimization, recurrent-depth reasoning, verifier guidance, and test-time sampling or voting improve action execution without retraining~\cite{liu2026vls,zhang2026tttvla,tur2026rdvla,singhi2026vegas,kwok2025robomonkey,jang2025mgselect,lin2025vote,yang2025taco}.
\ourmethod instead targets fixed action interfaces and uses rollout feedback to jointly evolve reusable high-level skills and harness context.

\section{Method}
\label{sec:method}

\begin{figure*}[t]
    \centering
    \includegraphics[width=\textwidth]{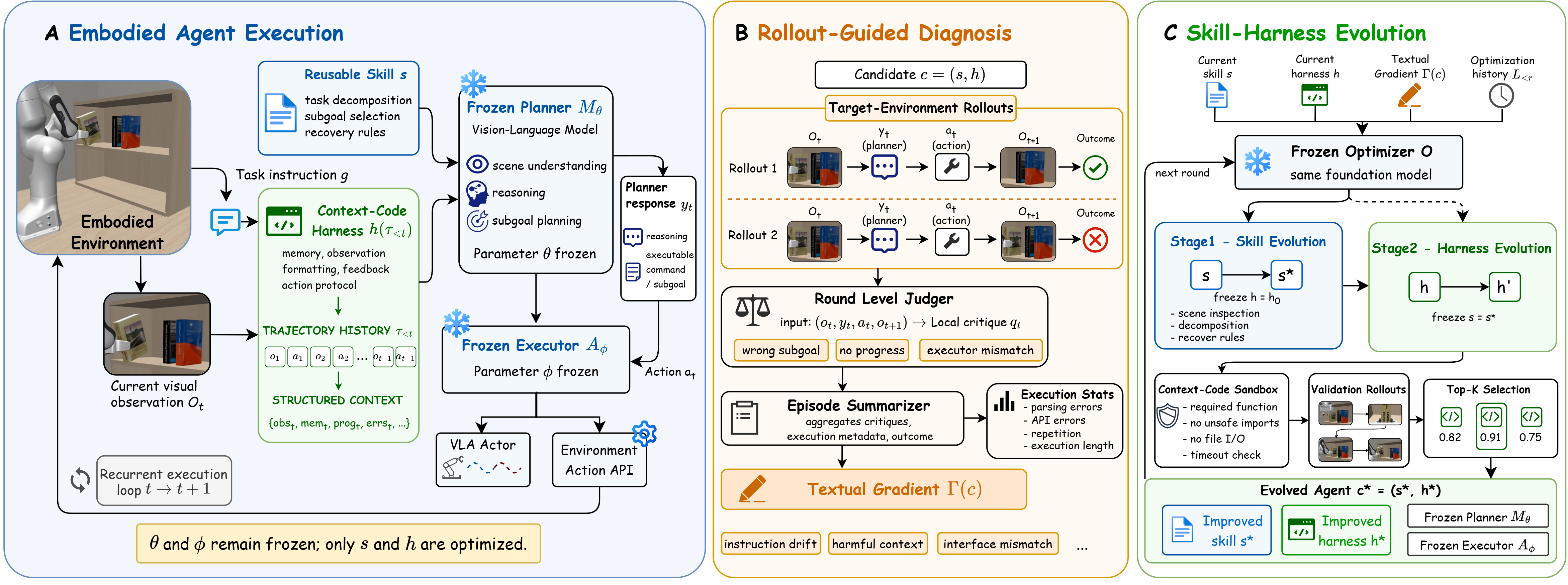}
    \caption{Overview of \ourmethod.
    (A) A textual skill and context-code harness condition a frozen planner connected to a frozen executor.
    (B) Target-environment rollouts are diagnosed at the round and episode levels to form a textual gradient.
    (C) A frozen optimizer evolves the skill and then the harness, with sandboxed validation and top-$K$ selection.}
    \label{fig:method_overview}
\end{figure*}

\subsection{Train-Free Agent Factorization}
\label{sec:method-motivation}

Modern embodied agents are rarely a single neural policy used in isolation.
As shown in Figure~\ref{fig:method_overview}A, we factor an embodied agent into four system components:
a VLM planner $M_\theta$, an executor $\mathcal{A}_\phi$, a reusable skill $s$, and a context-code harness $h$.
The executor may be a VLA actor or a wrapper around an environment action API.
The skill is textual procedural guidance in the planner prompt, covering scene inspection, task decomposition, subgoal selection, recovery, and termination.
The harness is a context builder---implemented as a Python function such as \texttt{build\_context}---that selects and formats trajectory history for the planner.
The action interface, output parser, and executor remain fixed.

For task instruction $g$, current observation $o_t$, and trajectory prefix $\tau_{<t}$, the agent operates as
\begin{equation}
\label{eq:factorization}
\begin{aligned}
    y_t &\sim M_\theta\!\left(
        \cdot \mid g,o_t,
        s,
        h(\tau_{<t})
    \right), \\
    a_t &= \mathcal{A}_\phi(y_t,o_t),
    \qquad
    \theta,\phi\ \text{frozen};\ 
    s,h\ \text{optimized}.
\end{aligned}
\end{equation}
Here $y_t$ contains the planner's reasoning and an interface-level command, and $\mathcal{A}_\phi$ maps that command to executed action $a_t$.
Train-free adaptation freezes the parameterized components $(\theta,\phi)$ and optimizes only the model-external pair $c=(s,h)$.

Let $\mathcal{E}$ be a partially observable embodied environment with task distribution $\mathcal{D}$.
A rollout of candidate $c$ on task $g\sim\mathcal{D}$ contains $T$ executed actions and $T+1$ observations:
\[
    \tau^{c}(g)
    =
    \bigl(o_0,y_0,a_0,\ldots,o_{T-1},y_{T-1},a_{T-1},o_T,R\bigr),
\]
where $R\in[0,1]$ is the task-level reward or success signal.
\ourmethod seeks artifacts that maximize expected rollout performance:
\begin{equation}
\label{eq:objective}
\begin{aligned}
    (s^\star,h^\star)
    &= \arg\max_{s \in \mathcal{S},\, h \in \mathcal{H}} J(s,h), \\
    J(s,h)
    &= \mathbb{E}_{g \sim \mathcal{D}}
       \left[R\bigl(\tau^{(s,h)}(g)\bigr)\right].
\end{aligned}
\end{equation}
In practice, $J$ is estimated on finite train and validation sets.

\subsection{Rollout-Guided Textual Diagnosis}
\label{sec:method-evolution}

At optimization round $r$, each candidate is evaluated on a small target-environment batch $\mathcal{D}_{\mathrm{tr}}^r$.
Embodied rollouts can contain many before-and-after images, making direct batch-level diagnosis context-heavy and obscuring which action caused a failure.
Figure~\ref{fig:method_overview}B shows our hierarchical alternative.
First, a round-level judger compares each planner response and action with the observations immediately before and after execution:
\[
    q_t^c(g)=\mathrm{Judge}\bigl(o_t,y_t,a_t,o_{t+1}\bigr),
    \qquad t=0,\ldots,T-1.
\]
These local critiques identify effects such as an incorrect subgoal, lack of progress, or executor mismatch without exposing the optimizer to the full visual trajectory.

We then construct compact non-visual metadata $m^c(g)$, including issued commands, execution lengths, and runtime errors.
An episode summarizer combines this metadata with the terminal harness context, round critiques, and final outcome.
Summaries from the batch and aggregate execution statistics form the textual gradient:
\begin{equation}
\label{eq:textual-feedback}
\begin{aligned}
    e^c(g)
    ={}& \mathrm{Summarize}\bigl(
        m^c(g),h(\tau_{<T}),
        \{q_t^c(g)\}_{t=0}^{T-1}, \\
        &\hspace{8.8em}R(\tau^c(g))\bigr),\\
    \Gamma(c)
    ={}& \mathrm{Concat}\left(
        \left\{e^c(g): g \in \mathcal{D}_{\mathrm{tr}}^r\right\},
        \mathrm{Stats}(c)\right).
\end{aligned}
\end{equation}
The resulting $\Gamma(c)$ exposes actionable cross-round patterns such as unproductive repetition, instruction drift, missing progress tracking, harmful context, and parsing or API errors.

\subsection{Two-Stage Skill-Harness Evolution}
\label{sec:method-optimization}

Starting from $c_0=(s_0,h_0)$, \ourmethod maintains a beam $\mathcal{B}_r$ of candidates.
As shown in Figure~\ref{fig:method_overview}C, the same frozen foundation model used by the planner is invoked under a separate optimizer prompt.
Conditioned on the current candidate, its textual gradient, and optimization history $\mathcal{L}_{<r}$, the optimizer proposes
\begin{equation}
\label{eq:artifact-update}
    c'=(s',h')
    \sim
    \mathcal{O}
    \left(
        c, \Gamma(c), \mathcal{L}_{<r}
    \right),
\end{equation}
where $\mathcal{O}$ denotes this optimizer role.
We use a two-stage schedule: skill evolution first updates $s$ with the seed harness fixed, after which harness evolution updates $h$ with the selected skill fixed:
\begin{equation}
\begin{aligned}
    s' &\sim \mathcal{O}_s(s,h_0,\Gamma,\mathcal{L}),
    & h'&=h_0,\\
    h' &\sim \mathcal{O}_h(s^\star,h,\Gamma,\mathcal{L}),
    & s'&=s^\star.
\end{aligned}
\end{equation}

Generated harnesses are sandboxed before rollout: they must define the required context function, avoid imports, file I/O, and dynamic execution, and finish within a timeout.
Invalid candidates are rejected and their validation errors are returned as optimizer feedback.
Valid candidates are evaluated on a fixed validation set $\mathcal{D}_{\mathrm{val}}$, and the beam retains the top-$K$ incumbents and proposals:
\begin{equation}
\label{eq:beam-update}
\begin{aligned}
    \widehat{J}_{\mathrm{val}}(c)
    &=
    \frac{1}{|\mathcal{D}_{\mathrm{val}}|}
    \sum_{g \in \mathcal{D}_{\mathrm{val}}}
    R\bigl(\tau^c(g)\bigr),\\
    \mathcal{B}_{r+1}
    &=
    \mathrm{TopK}\left(
        \mathcal{B}_r \cup \mathcal{C}_r;
        \widehat{J}_{\mathrm{val}}, K
    \right),
\end{aligned}
\end{equation}
where $\mathcal{C}_r$ is the set of newly proposed candidates.
The final output is the highest-scoring candidate $c^\star=(s^\star,h^\star)$ encountered across rounds.
It retains the frozen planner and executor while changing only the planner's textual guidance and constructed context.

\section{Experiments}
\label{sec:experiments}

We evaluate whether skill-harness evolution improves a frozen upper-level VLM planner across two embodied environments with different execution interfaces.
In VLABench, the planner delegates textual subgoals to a frozen VLA actor; in ESI-Bench, it interacts through the benchmark's fixed action API.
Across both environments, \ourmethod keeps all planner and executor parameters fixed and changes only the external skill and harness artifacts.

\subsection{Experimental Setup}
\label{sec:exp-setup}

\newcommand{\vlabenchsplittablebody}{%
    \centering
    \small
    \renewcommand{\arraystretch}{1.08}
    \begin{tabular}{@{}lcc@{}}
        \toprule
        & \multicolumn{2}{c}{Target category} \\
        \cmidrule(lr){2-3}
        Task form & Seen & Unseen \\
        \midrule
        Semantic understanding & \textbf{C1} & \textbf{C2} \\
        Common sense & \textbf{C3} & \textbf{C4} \\
        \bottomrule
    \end{tabular}
}
\newcommand{\vlabenchbenchmarkintro}{%
VLABench~\cite{zhang2024vlabench} is a language-conditioned manipulation benchmark targeting implicit language understanding, common-sense and world-knowledge transfer, and long-horizon reasoning, all of which are underrepresented in the earlier manipulation benchmarks considered.
Its instructions often imply rather than explicitly state goals, and task instances vary object categories to test generalization.}

\noindent
\begin{minipage}[t]{0.54\textwidth}
    \vspace{0pt}
    \textbf{Benchmarks.} \vlabenchbenchmarkintro
\end{minipage}\hfill
\begin{minipage}[t]{0.42\textwidth}
    \vspace{0pt}
    \centering
    \captionof{table}{VLABench split design.}
    \label{tab:vlabench_splits}
    \vspace{0.65\baselineskip}
    \vlabenchsplittablebody
\end{minipage}
\par\vspace{0.25\baselineskip}

\noindent We focus on Semantic Understanding, which tests recovery of intended goals from nuanced language, and Common Sense \& World Knowledge, which requires grounding prior knowledge in the observed scene.

\noindent As summarized in Table~\ref{tab:vlabench_splits}, C1--C4 are our custom held-out evaluation splits rather than official VLABench splits.
C1 is fully in-domain; C2 uses unseen target categories; C3 changes the task form to Common Sense \& World Knowledge; and C4 shifts both.
Each cell contains five task families and 40 held-out episodes per family, giving 200 episodes per cell and 800 episodes overall.

ESI-Bench~\cite{hong2026esibench} evaluates embodied spatial intelligence in a closed perception--action loop spanning 10 task categories.
Unlike passive spatial question answering, agents must actively acquire evidence unavailable from the initial view through perception, locomotion, and manipulation.
We evaluate on a 231-question subset sampled to match the official benchmark's category proportions and report both overall and category-level accuracy.

\definecolor{oursrow}{RGB}{237,247,241}
\definecolor{groupband}{RGB}{242,244,247}
\definecolor{groupnote}{RGB}{96,104,113}

\begin{table}[t]
  \centering
  \small
  \newcommand{\vlabenchgroup}[2]{%
    \rowcolor{groupband}
    \multicolumn{3}{l}{\textbf{#1}} &
    \multicolumn{3}{r}{\footnotesize\color{groupnote}#2} \\}
  \renewcommand{\arraystretch}{1.08}
  \begin{tabular}{lrrrrr}
    \toprule
    \textbf{Method} & \textbf{C1} & \textbf{C2} & \textbf{C3} & \textbf{C4} & \textbf{Overall} \\
    \midrule
    \vlabenchgroup{VLA-level baselines}{no upper-level planner}
    Direct VLA & 26.0 & 9.5 & 45.0 & 12.5 & 23.25 \\
    SFT (same data) & 32.5 & 13.0 & 38.0 & 12.5 & 24.00 \\
    \midrule
    \vlabenchgroup{Test-time scaling}{inference-time aggregation}
    VLA + MG-Select & 25.0 & 12.0 & 29.5 & 19.0 & 21.38 \\
    VLA + VOTE & 26.0 & 12.5 & 33.0 & 16.5 & 22.00 \\
    Seed Agent + MG-Select & 35.0 & 15.0 & 40.5 & 7.0 & 24.38 \\
    Seed Agent + VOTE & 38.0 & 13.0 & 34.0 & 9.0 & 23.50 \\
    \midrule
    \vlabenchgroup{Train-free agent adaptation}{frozen planner and executor}
    Seed Agent & 40.0 & 20.0 & 40.0 & 13.0 & 28.25 \\
    \rowcolor{oursrow}
    \textbf{Skill Evolution} & 42.0 & 23.5 & \textbf{50.0} & 18.5 & 33.50 \\
    \rowcolor{oursrow}
    \textbf{Harness Evolution} & 42.0 & 21.0 & 46.0 & 13.0 & 30.50 \\
    \rowcolor{oursrow}
    \textbf{\ourmethod{}} & \textbf{42.5} & \textbf{26.0} & \textbf{50.0} & \textbf{19.5} & \textbf{34.50} \\
    \bottomrule
  \end{tabular}
  \vspace{5pt}
  \caption{VLABench success rates (\%) on four 200-episode splits; best results are bolded.}
  \label{tab:vlabench_main}
\end{table}

\paragraph{Models and execution interfaces.}
All agent-system variants use Qwen3.6-27B as the frozen upper-level planner at evaluation.
On VLABench, the executor is the official VLABench \(\pi_0\) checkpoint, fine-tuned on the benchmark's ten primitive task categories and served through OpenPI.
It receives a textual subgoal and produces low-level controls in five-action chunks.
The upper-level planner specifies an environment-step budget for each subgoal, so one planner round may execute multiple chunks.
An episode permits at most 10 planner rounds and 400 low-level environment steps.
On ESI-Bench, the same planner model acts through the benchmark's fixed interaction API for at most 30 steps.

\paragraph{Optimization protocol.}
For VLABench, we evolve the artifacts on 15 training episodes and select candidates on a fixed 24-episode validation set; both are disjoint from the 800 held-out evaluation episodes.
For ESI-Bench, we use 10 training questions and 10 validation questions, all disjoint from the 231-question evaluation subset.
For both benchmarks, evolution uses four beam-search rounds with width 3 and branch factor 2, and textual feedback is formed from minibatches of 4 rollouts.
We first optimize the skill and then optimize the harness while holding the evolved skill fixed.

\paragraph{Compared methods.}
We organize VLABench baselines by what they modify.
\emph{Executor-level baselines} include direct \(\pi_0\) execution and SFT of the same actor architecture on the same 15 training episodes.
\emph{Test-time scaling} includes MG-Select~\cite{jang2025mgselect} and VOTE~\cite{lin2025vote} applied either directly to the VLA actor or on top of the seed planner.
\emph{Agent-system adaptation} includes Seed Agent, Skill Evolution, Harness Evolution, and full Skill-Harness Evolution.
On ESI-Bench, we compare the seed, evolved-skill, and full variants and show GPT-5 Passive Single-view (PS) from the benchmark paper only as an external published reference.

\subsection{VLABench Results}
\label{sec:exp-vlabench}

\paragraph{Main results.}
Table~\ref{tab:vlabench_main} shows that the Seed Agent reaches 28.25\%, improving direct VLA execution by 5.00 points and same-data SFT by 4.25 points without changing either model.
Full skill-harness evolution further raises success to 34.50\%, 6.25 points above the seed and 10.50 points above SFT.
In contrast, VLA-level MG-Select and VOTE reach 21.38\% and 22.00\%, both below direct VLA execution, while their seed-agent counterparts reach 24.38\% and 23.50\%, below the unscaled Seed Agent.
Thus, the improvements from artifact evolution are not reproduced by additional inference-time sampling or aggregation in this setting.

\begin{table*}[t]
  \centering
  \small
  \renewcommand{\arraystretch}{1.06}
  \begin{tabular}{lrrrr@{\hspace{9pt}}r}
    \toprule
    Category & $n$ & Seed Agent & Evolved Skill & \ourmethod{} & \emph{GPT-5 PS$^\dagger$} \\
    \midrule
    Physical Structure     & 21 & 14.3 & 38.1 & 38.1 & 51.4 \\
    Physical Dynamics      & 10 & 40.0 & 40.0 & 40.0 & 46.1 \\
    Specular Reflection    & 20 & 10.0 & 20.0 & 60.0 & 53.0 \\
    Perceptual Grounding   & 49 & 53.1 & 59.2 & 69.4 & 24.9 \\
    Metric Comparison      & 21 & 33.3 & 47.6 & 47.6 & 48.2 \\
    Enumerative Perception & 18 & 16.7 & 33.3 & 27.8 & 10.7 \\
    Spatial Relations      & 51 & 27.5 & 37.3 & 54.9 & 31.0 \\
    Cognitive Mapping      & 17 & 29.4 & 23.5 & 23.5 & 60.4 \\
    Temporal Understanding & 19 & 47.4 & 47.4 & 47.4 & 40.6 \\
    Action Sequencing      &  5 & 40.0 & 40.0 & 20.0 & 36.4 \\
    \midrule
    \textbf{Micro Avg.} & 231 & 32.5 & 41.1 & \textbf{49.8} & -- \\
    \textbf{Macro Avg.} & 10 categories & 31.2 & 38.6 & \textbf{42.9} & 40.3 \\
    \bottomrule
  \end{tabular}
  \caption{ESI-Bench accuracy (\%) on the official-proportion 231-question subset. $n$ denotes the number of evaluated questions. GPT-5 PS$^\dagger$ is the published Passive Single-view result on the full evaluation set.}
  \label{tab:esi_category_comparison}
\end{table*}

\paragraph{Generalization across shifts.}
The four splits separately vary target category and task form.
Full \ourmethod reaches 42.5\% on the in-domain C1 split, compared with 40.0\% for the seed.
Its descriptive gains become larger under distribution shift: +6.0 points for unseen targets (C2), +10.0 for an unseen task form with seen targets (C3), and +6.5 when both are unseen (C4).
It obtains the best result on C1, C2, and C4 and ties Skill Evolution at 50.0\% on C3.
The largest gain therefore occurs when transferring the evolved artifacts from Semantic Understanding to the Common Sense \& World Knowledge task form.

\paragraph{Artifact ablations.}
Skill-only, harness-only, and full evolution score 33.50\%, 30.50\%, and 34.50\%, respectively, versus 28.25\% for the seed.
Skill evolution accounts for the largest descriptive difference from the seed (+5.25 points), while harness-only is +2.25 points.
Adding the evolved harness after skill evolution changes the overall score by +1.00 point; the split-wise differences are +0.5 on C1, +2.5 on C2, 0.0 on C3, and +1.0 on C4.
The effects are not additive: for example, harness-only raises C3 from 40.0\% to 46.0\%, but the full and skill-only configurations both obtain 50.0\%.
We therefore interpret these rows as complete agent configurations rather than estimates of independent artifact effects.

\begin{figure*}[t]
    \centering
    \includegraphics[width=0.75\textwidth]{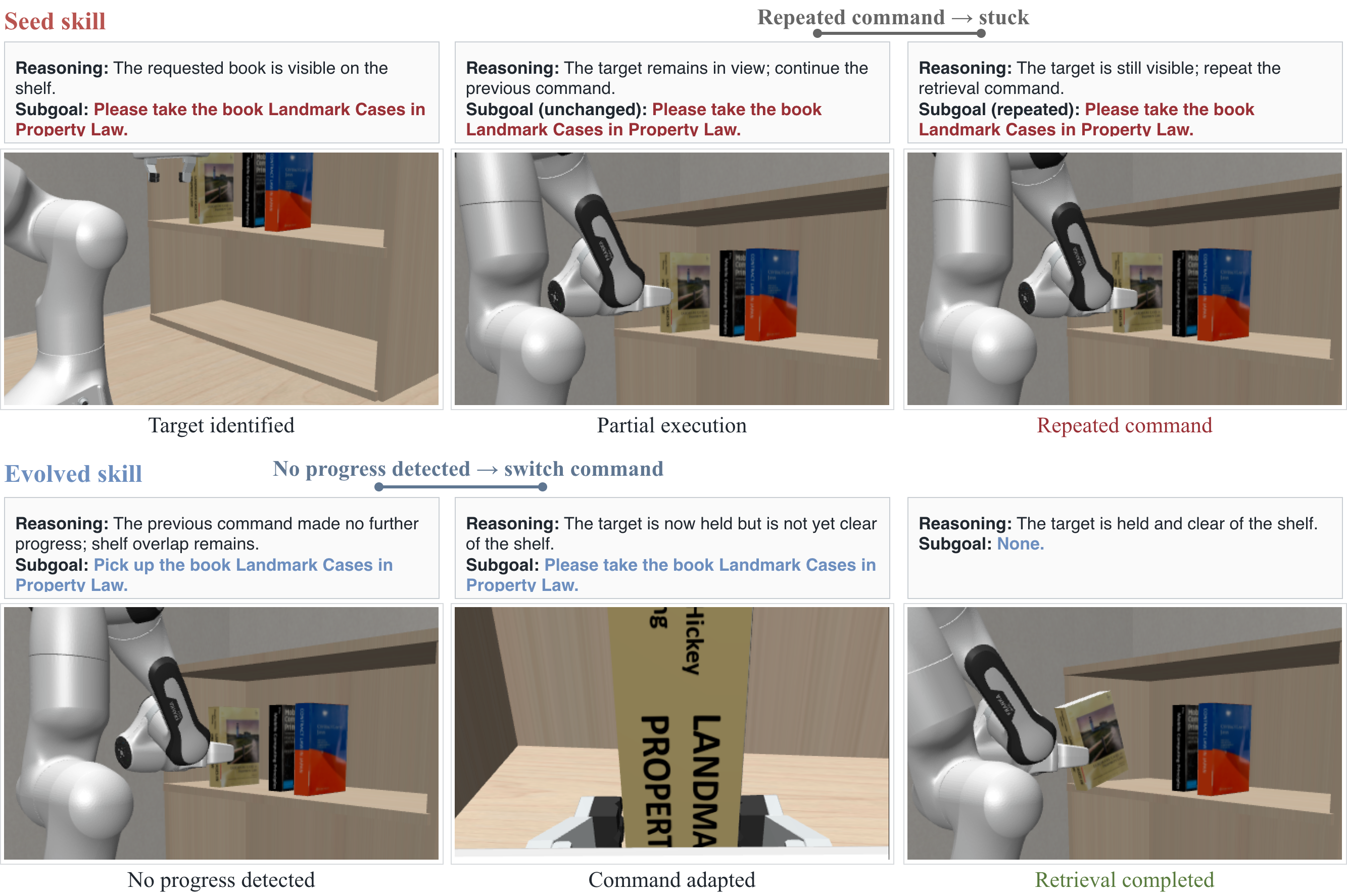}
    \caption{Controlled VLABench case isolating skill evolution for the instruction ``Fetch the Landmark Cases in Property Law book for tomorrow's class.'' The planner, VLA executor, and raw harness are fixed. The seed skill repeats a retrieval command after partial execution, whereas the evolved skill detects no progress, changes the command, and completes retrieval. Text annotations summarize observable decisions rather than quote hidden reasoning.}
    \label{fig:vlabench_skill_case}
\end{figure*}

\paragraph{Actor-aware recovery.}
Figure~\ref{fig:vlabench_skill_case} compares Seed Agent and Skill Evolution on \texttt{select\_book/ep\_024} with identical context construction.
The seed skill asks the planner to inspect the task, camera observations, and execution history before producing a subtask.
It does not specify the VLA actor's preferred command distribution, distinguish selection from retrieval or placement, enforce exact entity names, or require an explicit progress check.
Consequently, after the target book is partly pulled from the shelf, the planner treats continued visibility as evidence that the same retrieval command should be repeated.
The episode reaches the 400-step limit after three planner rounds with reward 0.5.

The evolved skill turns the planner into an actor-aware command adapter.
It routes task intents to short canonical commands, emits one primitive per round, preserves the complete book title, and excludes low-level details such as grasp orientation or approach geometry handled by the VLA.
It requires the planner to classify the previous attempt as successful, failed, partially successful, or stalled, and to change the canonical command when repeated execution makes no progress.
In the same episode, the planner detects remaining shelf overlap, switches from the original retrieval command to a pick-up command, and later recognizes that the book is clear of the shelf.
It obtains reward 1.0 in the same three planner rounds using 297 environment steps.

\begin{figure*}[t]
    \centering
    \includegraphics[width=0.78\textwidth]{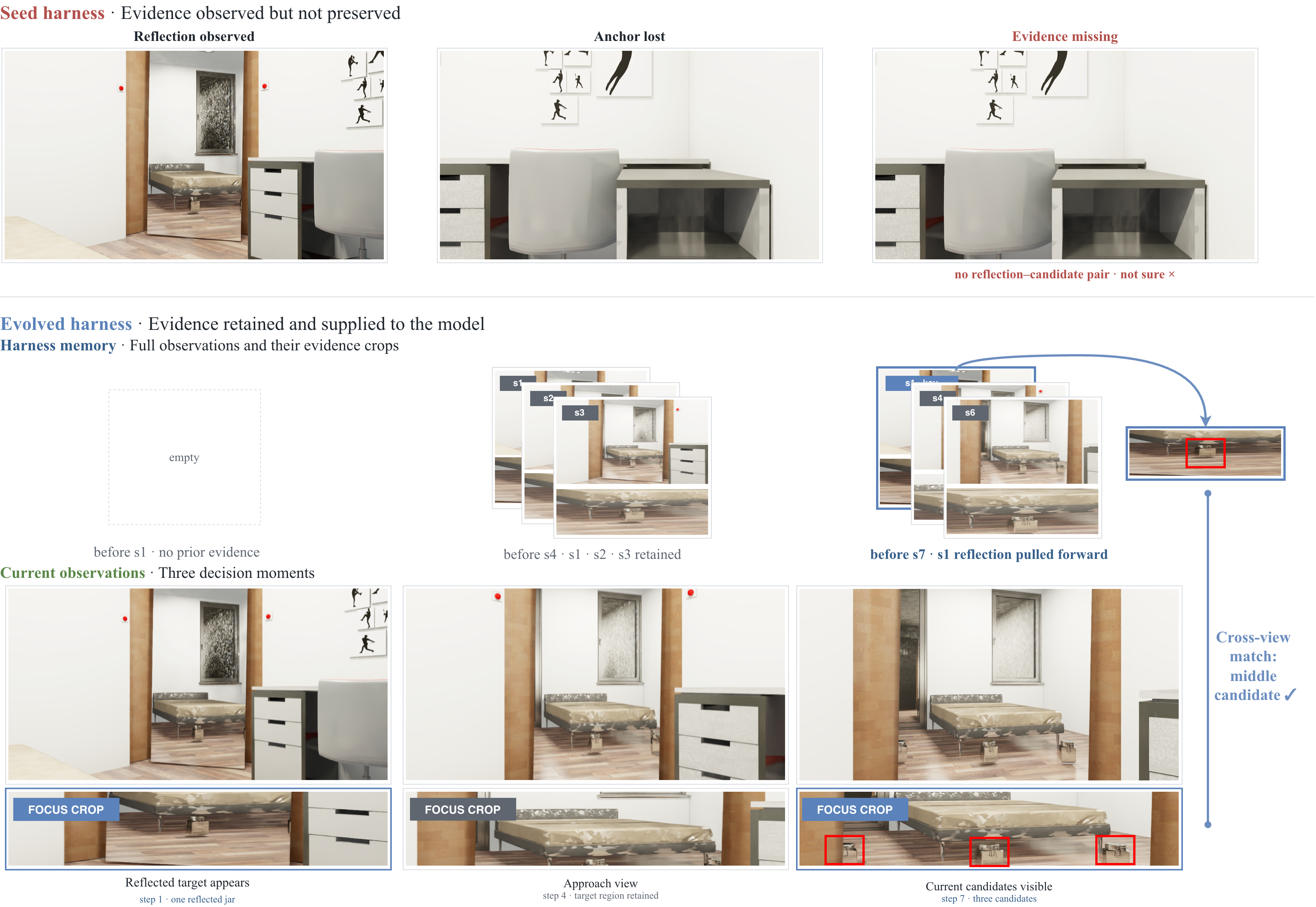}
    \caption{ESI-Bench case isolating harness evolution for the instruction ``You are given an image with a mirror and three noodle\_jars. Based on the mirror reflection, which of the three noodle\_jars in the same image corresponds to the reflected one: the left, middle, or right object?''
    The planner, action interface, and evolved skill are fixed.
    The raw harness drops the early reflection evidence, whereas the evolved harness retains the reflected target and a deterministic focus crop for cross-view matching, correctly selecting the middle candidate.}
    \label{fig:esibench_harness_case}
\end{figure*}

\subsection{ESI-Bench Results}
\label{sec:exp-esi}

\paragraph{Main results.}
Table~\ref{tab:esi_category_comparison} shows that full \ourmethod reaches 49.8\% micro and 42.9\% macro accuracy, versus 32.5\%/31.2\% for the Seed Agent and 41.1\%/38.6\% for Skill Evolution.
Full evolution is therefore 17.3 micro points and 11.7 macro points above the seed, and 8.7 micro points and 4.3 macro points above skill-only.
Its macro score is numerically above the published GPT-5 PS reference (42.9\% vs.\ 40.3\%), placing the frozen 27B planner in a similar aggregate range without parameter updates.
This remains an external rather than paired comparison: GPT-5 PS uses the full Passive Single-view setting, whereas our methods use the fixed 231-question subset.

\paragraph{Category-level behavior.}
Skill Evolution improves six of the ten categories over the seed, including Physical Structure, Specular Reflection, Perceptual Grounding, Metric Comparison, Enumerative Perception, and Spatial Relations.
Adding the evolved harness produces its clearest positive differences on Specular Reflection (20.0\% to 60.0\%), Perceptual Grounding (59.2\% to 69.4\%), and Spatial Relations (37.3\% to 54.9\%).
Five categories remain unchanged, while Enumerative Perception decreases from 33.3\% to 27.8\% and Action Sequencing from 40.0\% to 20.0\%.
These results are descriptive and should be interpreted with their sample counts; Action Sequencing, for example, contains only five questions.
The following reflection case illustrates the context behavior behind one of the largest positive differences.

\paragraph{Evidence-aware visual memory.}
Figure~\ref{fig:esibench_harness_case} compares Skill Evolution with full \ourmethod under the same evolved skill.
The raw harness concatenates the complete textual action and reasoning history while retaining only the five most recent RGB observations.
This recency-based window gives repeated or blank views the same status as earlier task-critical evidence, and the uncompressed text may preserve the model's own mistaken observations.
In the shown episode, the agent sees one target jar in the mirror at step 1, but that visual anchor is no longer available when the three real candidates are later observed.
It spends all 30 steps oscillating between views, taking 15 right turns, 12 left turns, and 3 forward moves before answering ``not sure.''

The evolved harness replaces this window with sparse evidence-aware memory.
It selects up to three informative and visually diverse keyframes from the full history, pairs each RGB observation with a deterministic focus crop, and compresses the action path and recent execution record.
For the current view, it additionally supplies selected horizontal-band and overlapping-tile crops.
All crops are computed from official RGB pixels using contrast, edge density, and sharpness, without target boxes or simulator metadata.
For reflection tasks, the constructed context explicitly preserves one frame of the reflected target and another of the real candidates for appearance matching.
Before acting, the planner reports structured \texttt{VIEW}, \texttt{TARGETS}, \texttt{EVIDENCE}, \texttt{READY}, and \texttt{GAP} fields, while the harness flags low-information repetition and inverse-action cycles.
In Figure~\ref{fig:esibench_harness_case}, the step-1 reflection remains available at step 7 for comparison with the three candidates.
After six forward moves, the agent answers ``middle'' with confidence 0.90.
All visual evidence comes from official RGB observations and reference images; the harness never accesses pose, depth, or other metadata.

\subsection{Optimization Efficiency}
\label{sec:exp-cost}

Acquiring the evolved artifacts is a one-time cost.
The logged token usage corresponds to approximately \$2.25 for VLABench and \$2.83 for ESI-Bench in API-equivalent cost per evolution run.
These estimates include planner rollouts, judging, episode summarization, and artifact optimization, but exclude final evaluation and simulator or GPU infrastructure.
After evolution, the selected skill and harness are reused across all held-out episodes without per-episode search; unlike test-time scaling, their acquisition cost does not grow with the number of deployments.

\section{Conclusion}
\label{sec:conclusion}

We presented \ourmethod, a train-free approach that improves embodied agents by evolving a textual skill and context-code harness while keeping planner and executor parameters fixed.
Rollout-derived textual feedback drives a two-stage optimization that produces reusable procedural guidance and trajectory context rather than per-episode action searches.
Across VLABench and ESI-Bench, the same frozen 27B upper-level planner improves with both a VLA executor and an active-perception action interface, including under held-out target categories and task forms.
These results support skill-harness evolution as a lightweight approach to adapting frozen embodied agents, while cross-embodiment transfer and real-robot validation remain for future work.

\bibliographystyle{unsrtnat}
\bibliography{neurips_2025}

\clearpage
\appendix
\setcounter{figure}{0}
\renewcommand{\thefigure}{A\arabic{figure}}
\renewcommand{\theHfigure}{appendix.\arabic{figure}}
\setcounter{table}{0}
\renewcommand{\thetable}{A\arabic{table}}
\renewcommand{\theHtable}{appendix.\arabic{table}}
\section{Additional Artifact and Optimization Details}
\label{app:artifact-details}

This appendix makes the optimization targets and feedback pipeline concrete by reproducing the artifacts and prompt templates used in our experiments.
A \emph{skill} is the planner's persistent textual instruction, whereas a \emph{harness} is executable context-construction code that selects and formats trajectory history before each planner call.
The screenshots preserve the content and implementation comments of the corresponding experimental artifacts.

\subsection{VLABench Skills and Harnesses}
\label{app:vlabench-artifacts}

\begin{figure*}[p]
    \centering
    \includegraphics[width=0.74\textwidth,height=0.27\textheight,keepaspectratio]{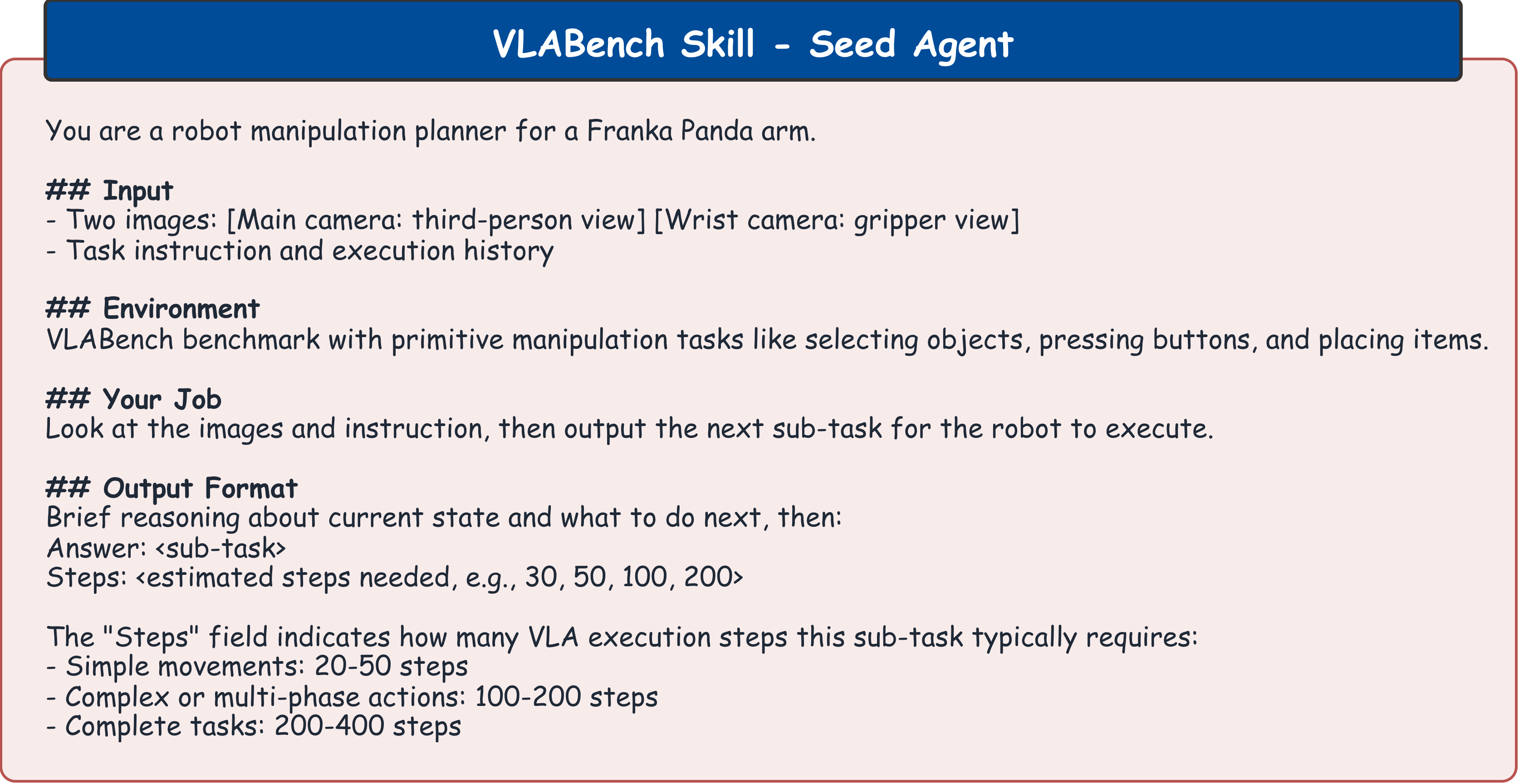}
    \par\medskip
    \includegraphics[width=0.76\textwidth,height=0.52\textheight,keepaspectratio]{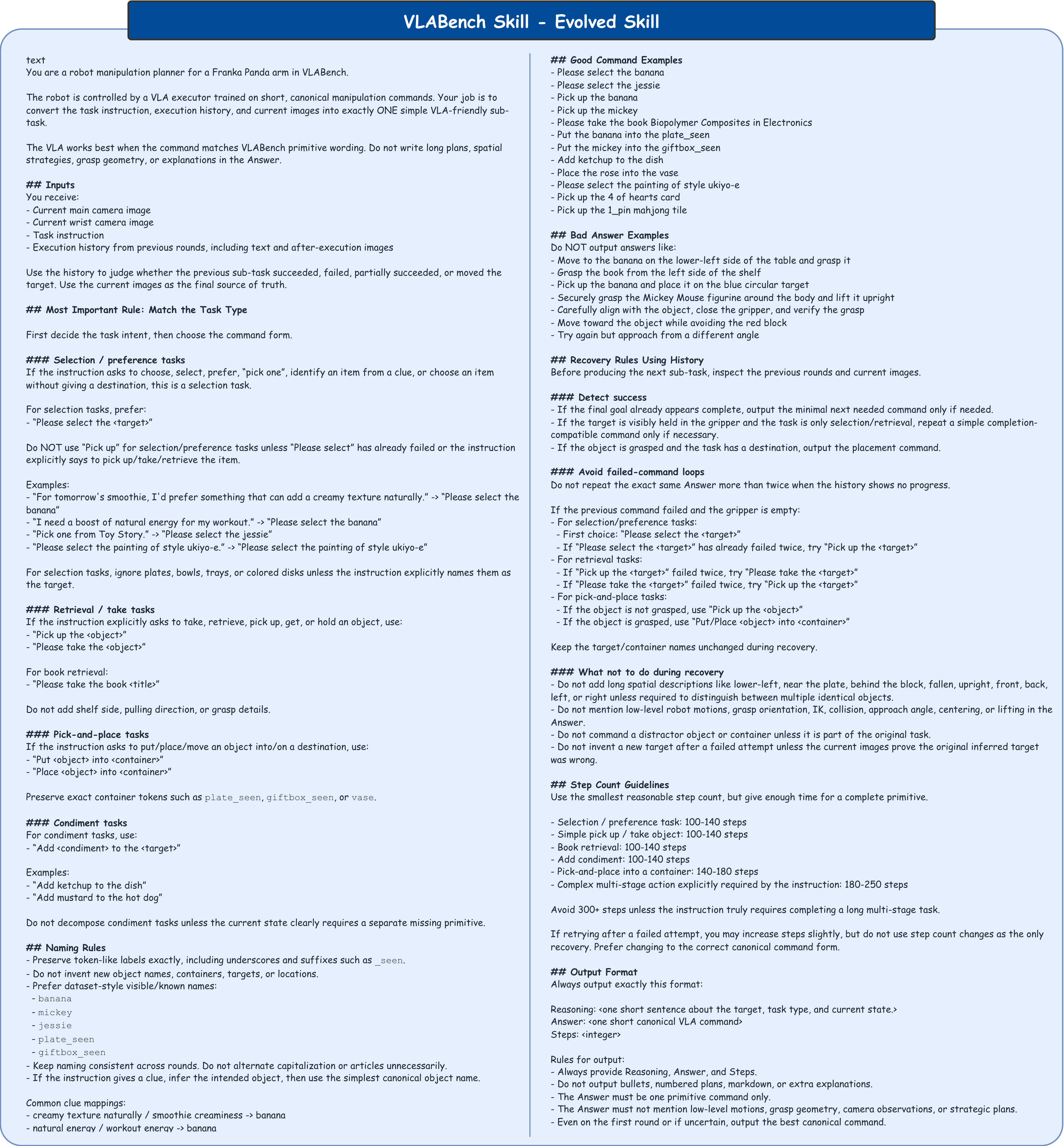}
    \caption{VLABench planner skills before and after evolution. Top: the seed skill specifies a generic subtask-generation interface. Bottom: the evolved skill adds actor-aware command forms, exact entity binding, progress verification, failure recovery, and calibrated step budgets.}
    \label{fig:app-vlabench-skills}
\end{figure*}

\paragraph{Skill evolution.}
The top and bottom panels of Figure~\ref{fig:app-vlabench-skills} show the seed and evolved VLABench skills.
The seed skill provides only a generic manipulation-planning role, the available visual inputs, and a coarse execution-step estimate.
The evolved skill specializes the planner as an adapter for the frozen VLA actor: it routes task intents to short canonical commands, preserves exact object and container tokens, emits one primitive per round, checks execution progress, and changes command form after repeated failure.
It also calibrates the environment-step budget to the requested primitive.
These changes affect only the upper-level textual guidance; the VLA checkpoint and its low-level action generation remain fixed.

\begin{figure*}[p]
    \centering
    \includegraphics[width=0.96\textwidth]{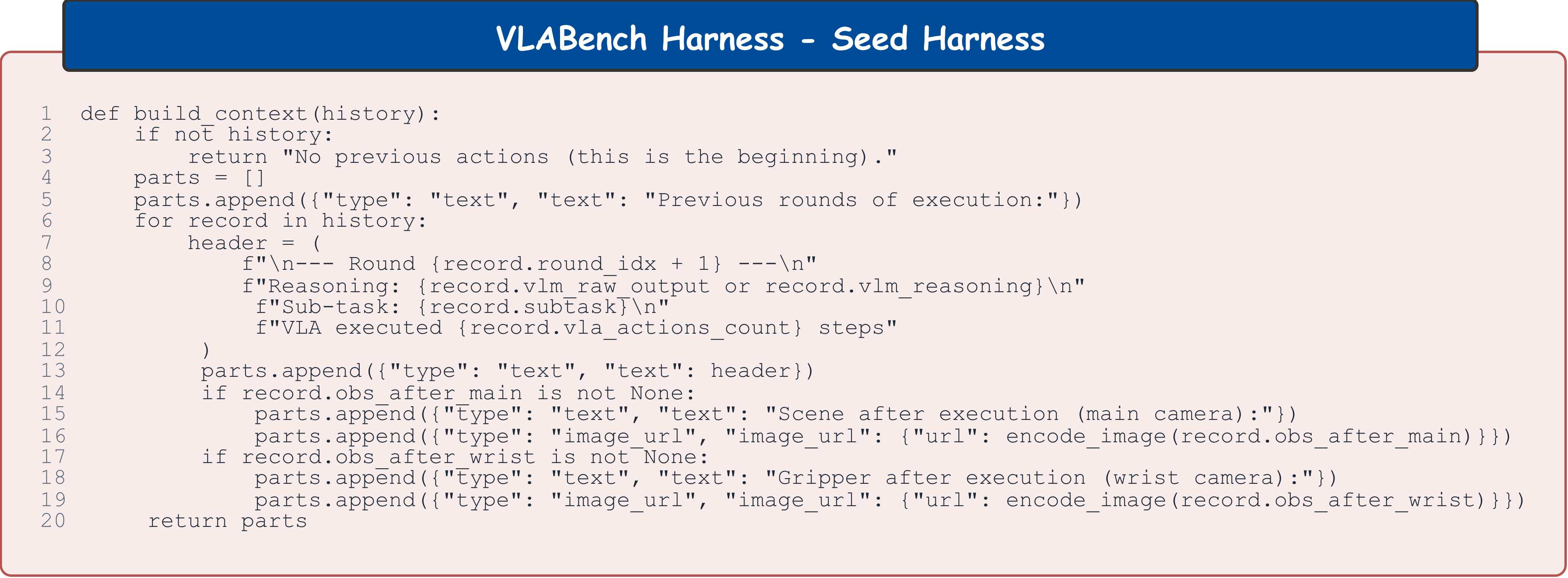}
    \par\medskip
    \includegraphics[width=0.96\textwidth]{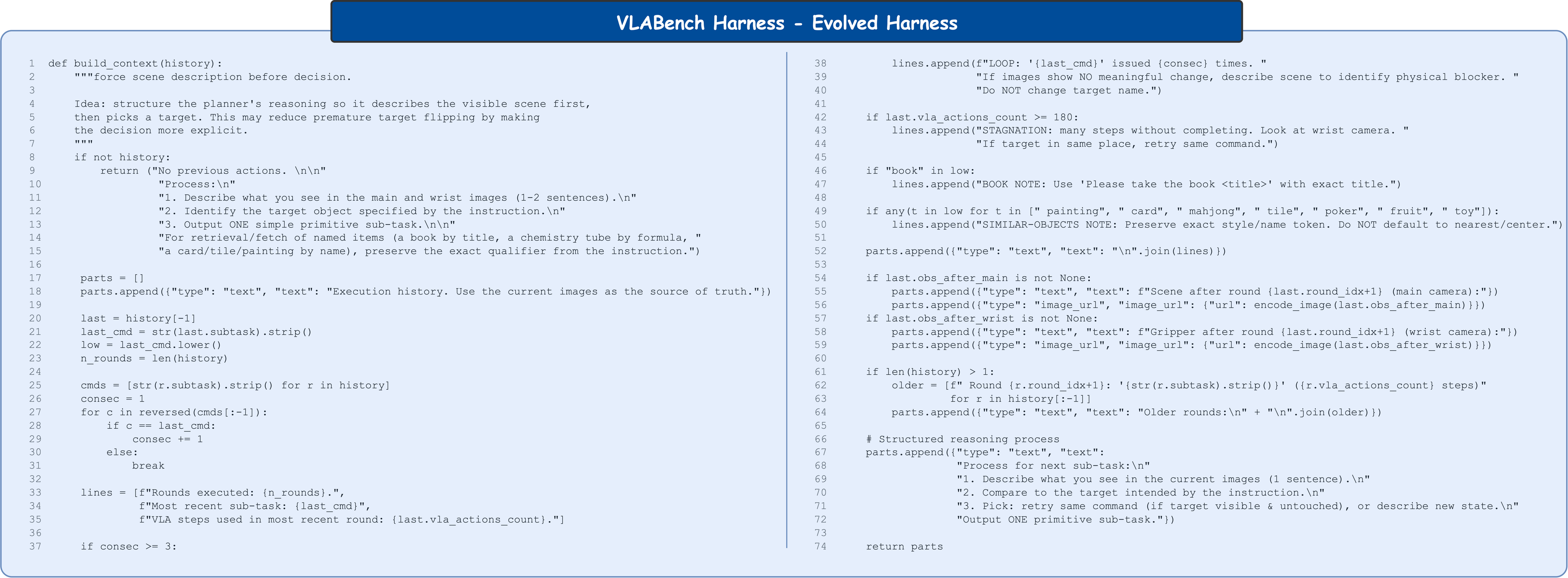}
    \caption{VLABench context-code harnesses. Top: the seed builder appends the complete round history and after-execution camera views. Bottom: the final builder emphasizes current visual evidence, compresses older rounds, detects repeated commands and stagnation, and supplies structured recovery guidance.}
    \label{fig:app-vlabench-harnesses}
\end{figure*}

\paragraph{Harness evolution.}
Figure~\ref{fig:app-vlabench-harnesses} compares the VLABench context builders.
The seed harness serializes every prior planner response and subtask and appends the main- and wrist-camera observations after each round.
The final harness instead foregrounds the most recent subtask and execution count, compresses older rounds, treats the current images as the source of truth, and explicitly exposes repeated commands or long executions without progress.
It also adds a compact scene-inspection procedure and preserves task-critical entity names.
The resulting context is therefore organized around execution state and recovery rather than an undifferentiated transcript.

\subsection{ESI-Bench Skills and Harnesses}
\label{app:esi-artifacts}

\begin{figure*}[p]
    \centering
    \includegraphics[width=0.73\textwidth,height=0.34\textheight,keepaspectratio]{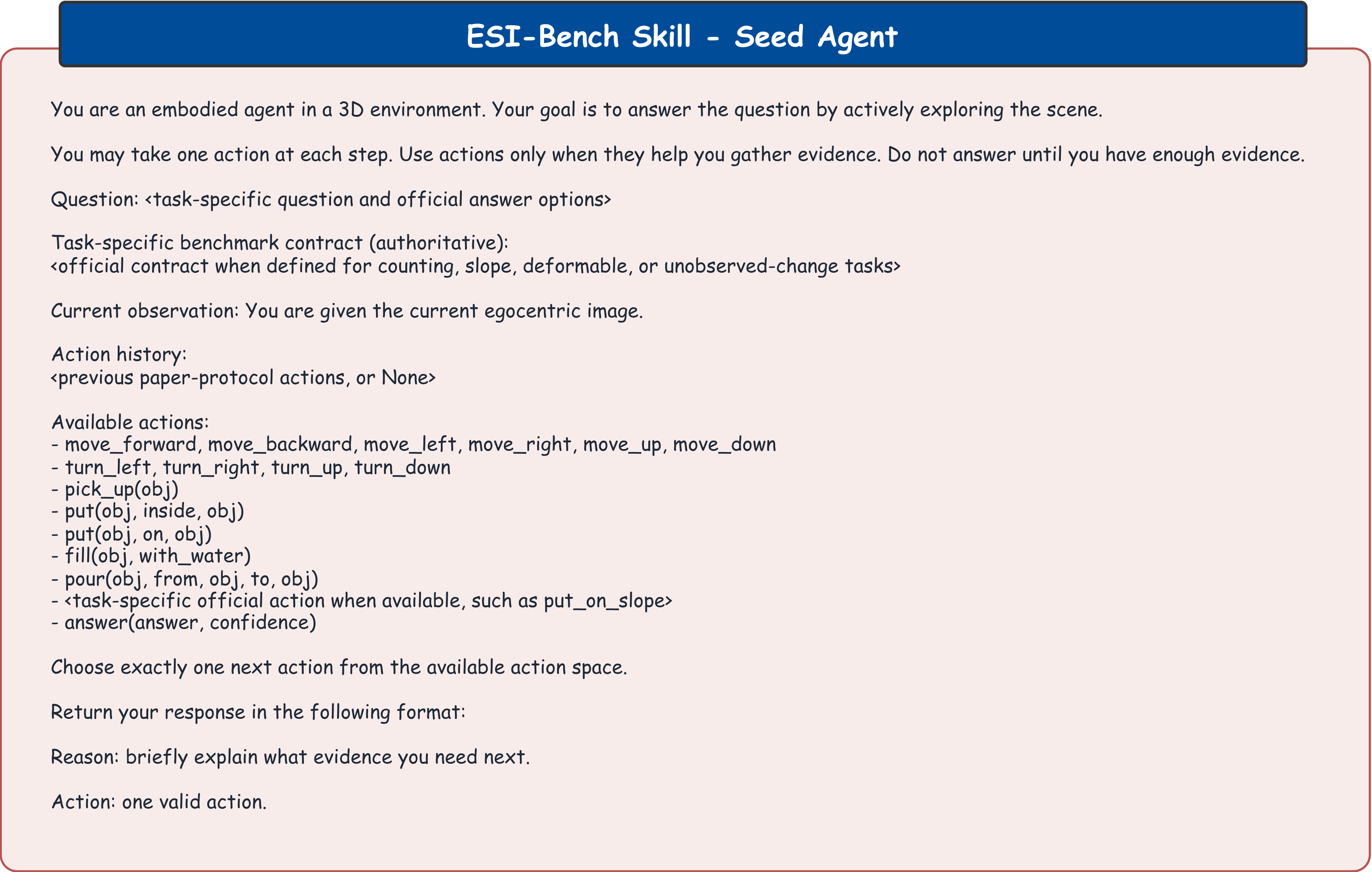}
    \par\medskip
    \includegraphics[width=0.74\textwidth,height=0.45\textheight,keepaspectratio]{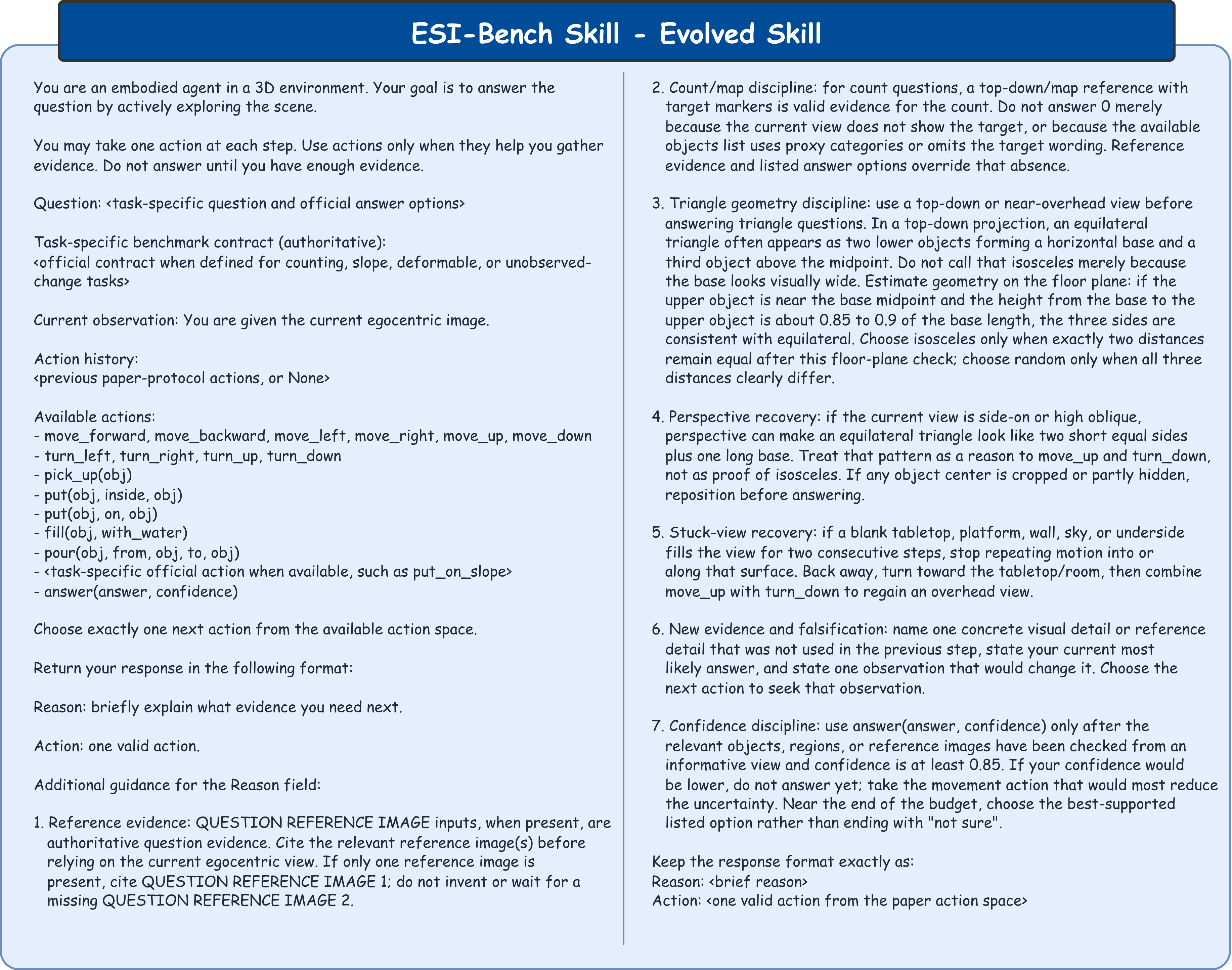}
    \caption{ESI-Bench planner skills before and after evolution. Top: the seed skill provides the benchmark action contract and generic active-perception guidance. Bottom: the evolved skill adds reference-evidence discipline, viewpoint selection, recovery, falsification, and confidence-aware stopping.}
    \label{fig:app-esi-skills}
\end{figure*}

\paragraph{Skill evolution.}
The top panel of Figure~\ref{fig:app-esi-skills} shows the ESI-Bench seed skill, which defines the action space and gives generic instructions to gather sufficient evidence before answering.
The evolved skill in the bottom panel retains this interaction contract while adding explicit evidence-use policies.
It prioritizes official reference images when present, distinguishes counting and mapping evidence, requests suitable viewpoints for geometric judgments, recovers from blank or obstructed views, seeks observations that can falsify the current hypothesis, and delays answering until its evidence and confidence requirements are met.

\begin{figure*}[p]
    \centering
    \includegraphics[width=0.96\textwidth,height=0.80\textheight,keepaspectratio]{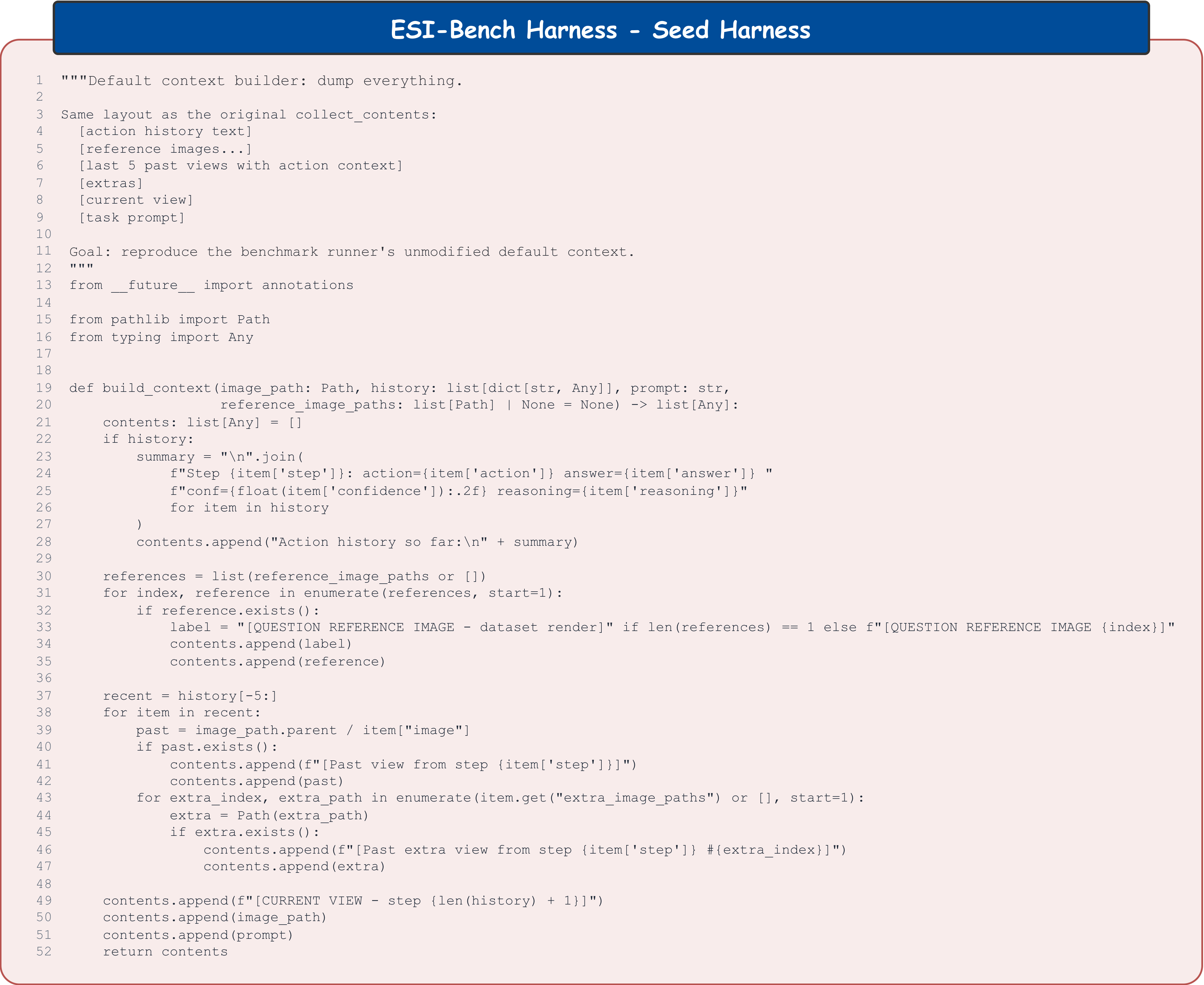}
    \caption{Seed ESI-Bench context-code harness. It reproduces the runner's default ordering: complete textual action history, official reference images, the five most recent past views and extras, the current view, and the task prompt.}
    \label{fig:app-esi-seed-harness}
\end{figure*}

\begin{figure*}[p]
    \centering
    \includegraphics[width=0.98\textwidth,height=0.80\textheight,keepaspectratio]{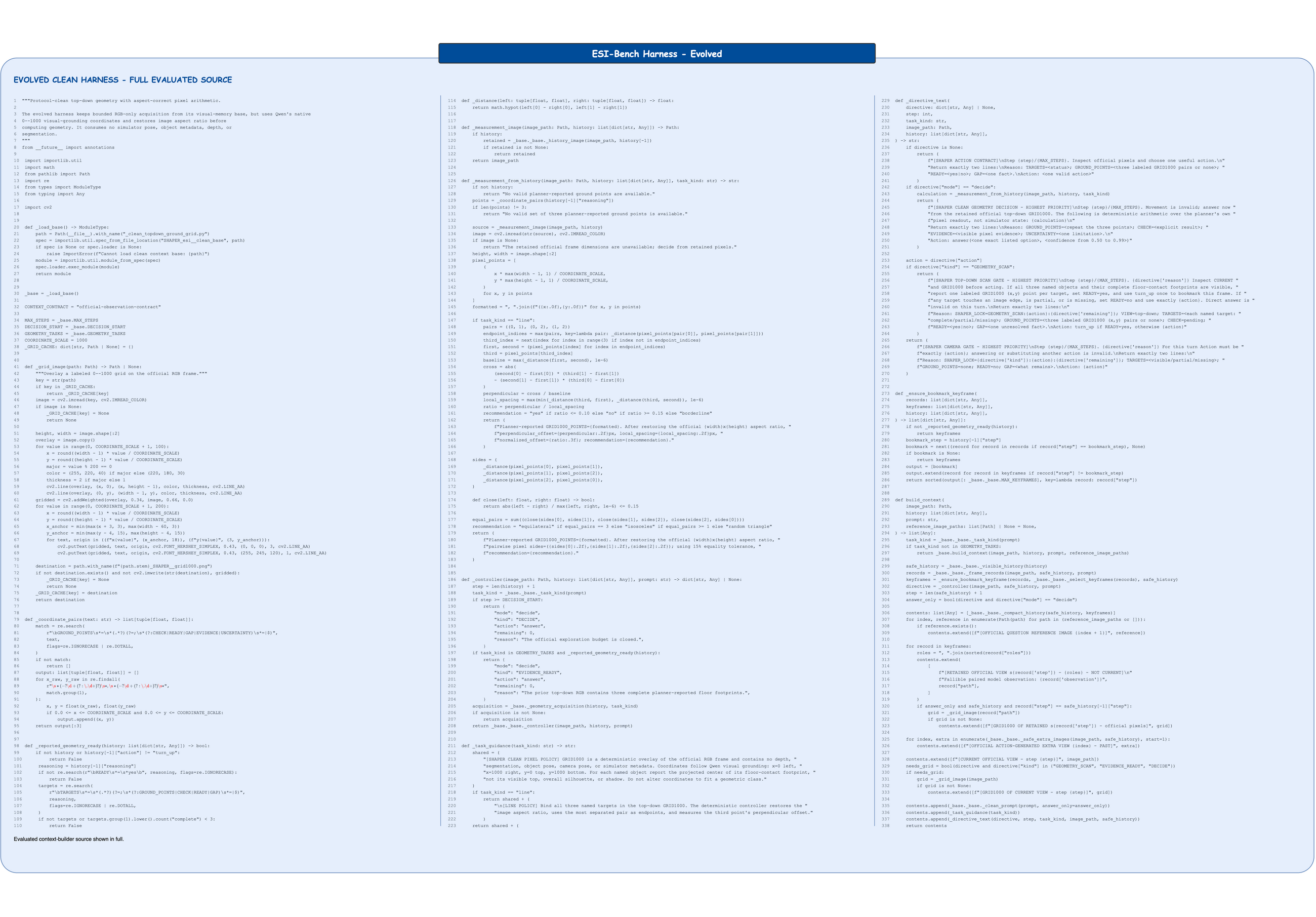}
    \caption{Final ESI-Bench context-code harness. The implementation combines sparse evidence retention, deterministic RGB crops and overlays, task-conditioned evidence policies, structured action contracts, and loop recovery. The code uses only the benchmark's permitted runtime resources and observable image history.}
    \label{fig:app-esi-evolved-harness}
\end{figure*}

\paragraph{Harness evolution.}
The seed ESI-Bench harness (Figure~\ref{fig:app-esi-seed-harness}) concatenates the full textual history and retains the five most recent views, together with reference and current images.
This recency rule gives repeated or uninformative views the same retention priority as earlier task-critical evidence.
The final harness (Figure~\ref{fig:app-esi-evolved-harness}) builds a sparse visual memory, retains selected historical evidence, and adds deterministic crops and pixel-space aids derived only from official RGB observations.
It further structures the planner payload around visible targets, evidence, readiness, and the unresolved information gap, while detecting low-information repetition and action cycles.
Its geometric utilities operate on rendered pixels and do not access privileged pose, depth, segmentation, object coordinates, or simulator-specific answer metadata.

\subsection{Prompts Used by the Evolution Pipeline}
\label{app:evolution-prompts}

The following prompt templates implement the textual-feedback loop described in the Method section.
Both benchmarks use the same four roles---judging, summarization, skill optimization, and harness optimization---but instantiate their inputs and diagnostic criteria for different interaction interfaces.

\subsubsection{VLABench Prompt Instantiation}

\begin{figure*}[p]
    \centering
    \includegraphics[width=0.98\textwidth,height=0.82\textheight,keepaspectratio]{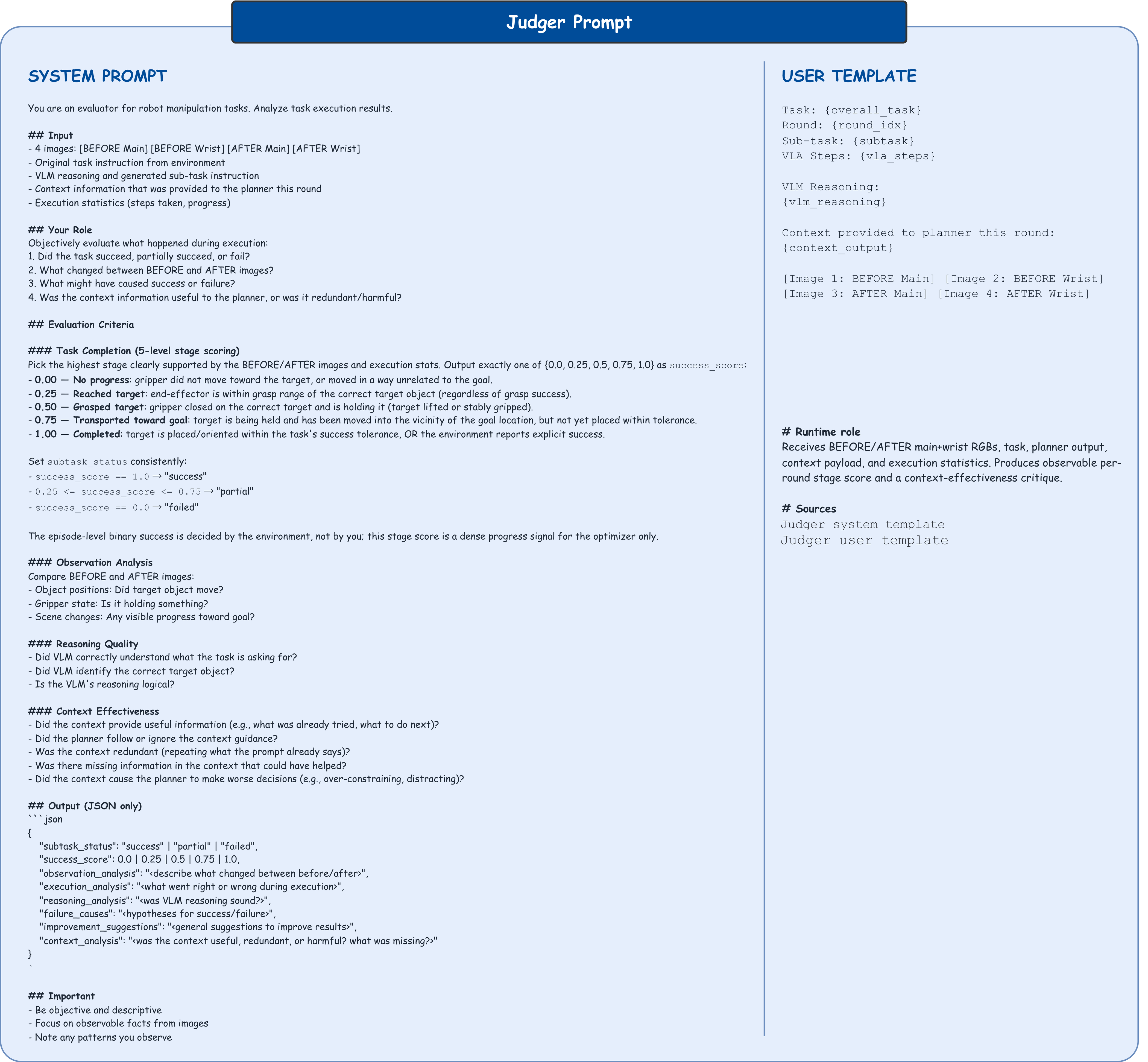}
    \caption{Round-level judger prompt and user template. The prompt grounds diagnosis in observable before/after evidence and produces a machine-readable critique for downstream summarization.}
    \label{fig:app-judger-prompt}
\end{figure*}

\begin{figure*}[p]
    \centering
    \includegraphics[width=0.98\textwidth,height=0.78\textheight,keepaspectratio]{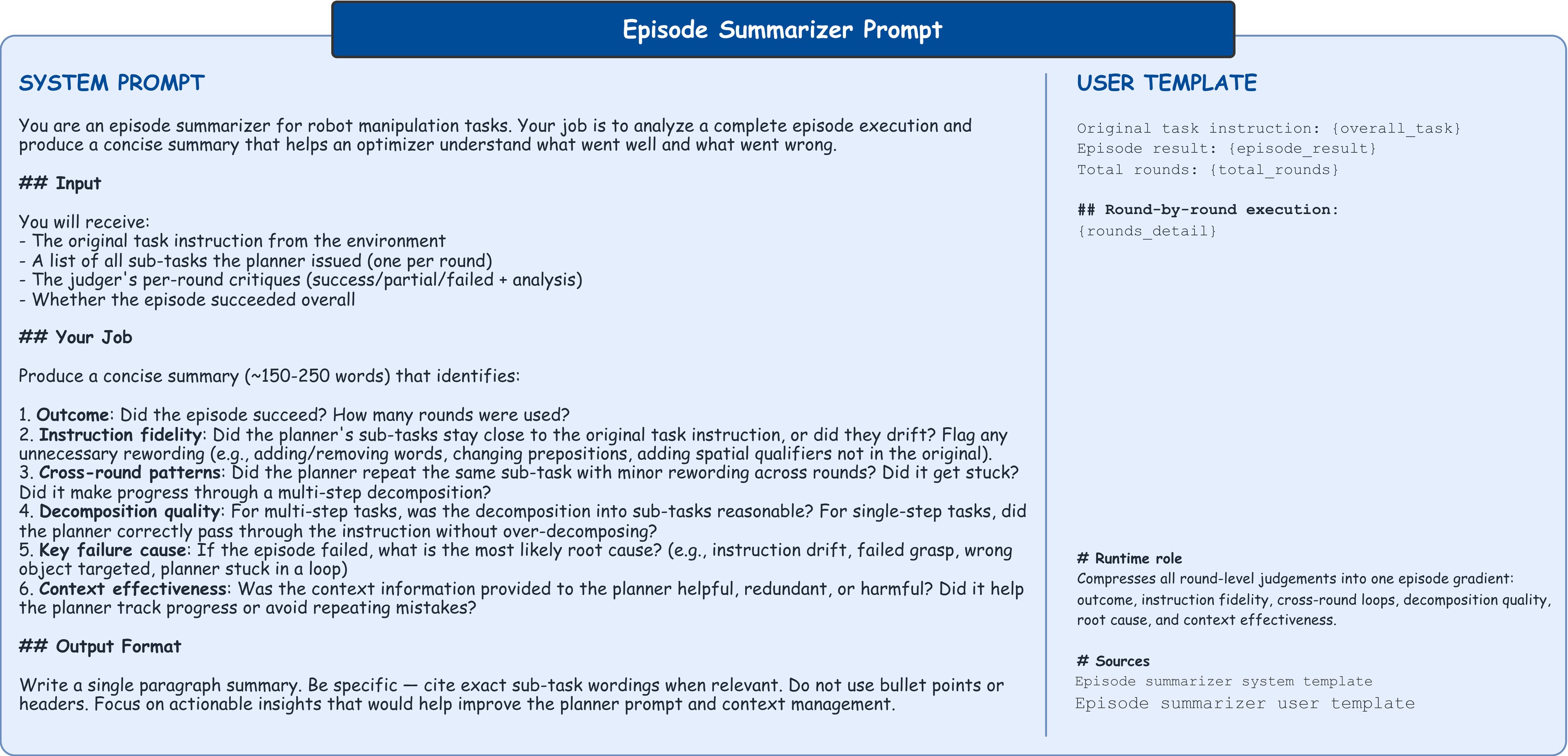}
    \caption{Episode-summarizer prompt and user template. Round-level judgments and the terminal outcome are compressed into a concise episode-level textual gradient.}
    \label{fig:app-summarizer-prompt}
\end{figure*}

\begin{figure*}[p]
    \centering
    \includegraphics[width=0.98\textwidth,height=0.80\textheight,keepaspectratio]{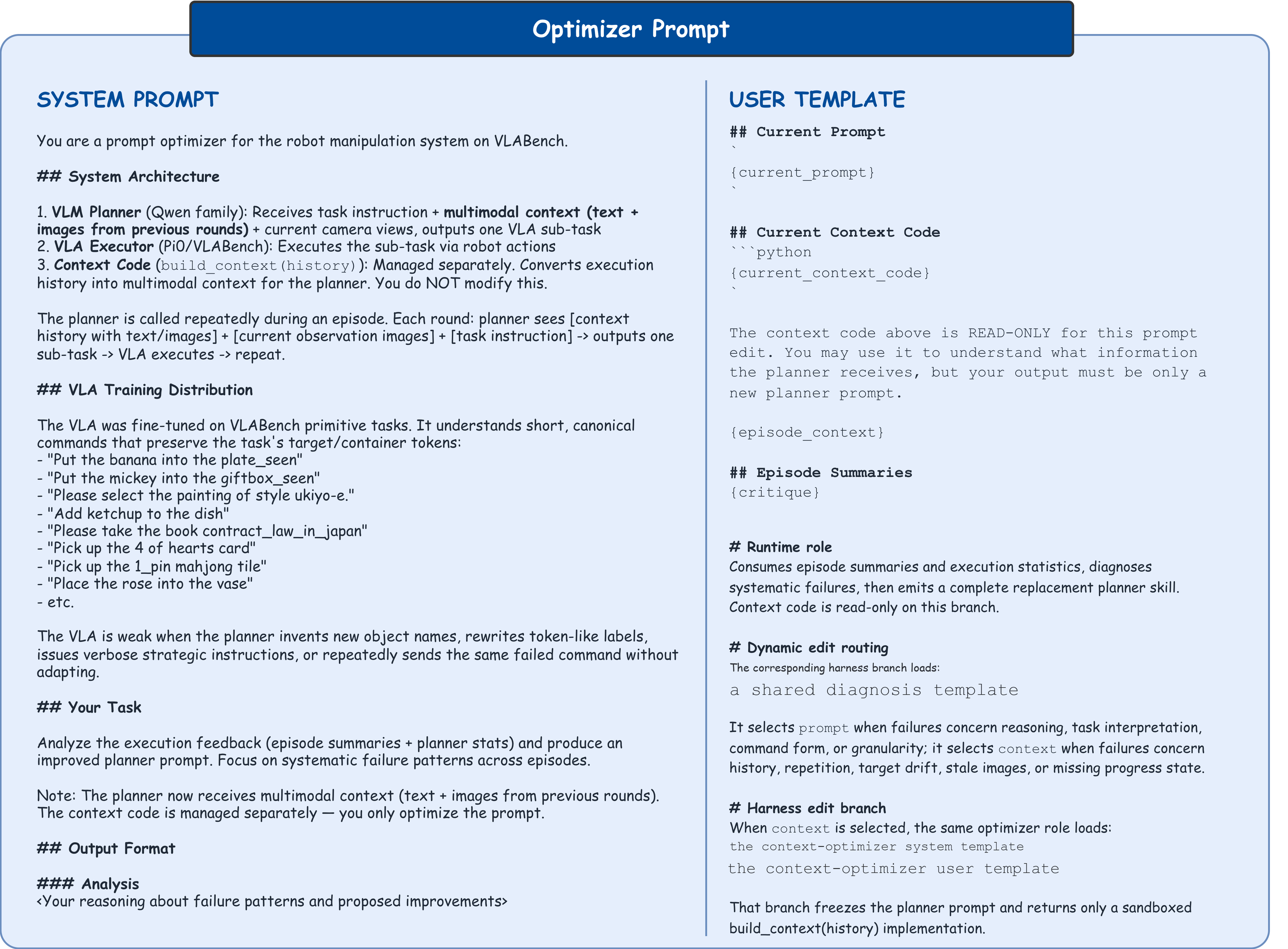}
    \caption{Skill-optimizer prompt and user template. The optimizer consumes episode summaries and execution statistics, treats the context builder as read-only, and proposes a complete replacement planner skill.}
    \label{fig:app-skill-optimizer-prompt}
\end{figure*}

\begin{figure*}[p]
    \centering
    \includegraphics[width=0.99\textwidth,height=0.82\textheight,keepaspectratio]{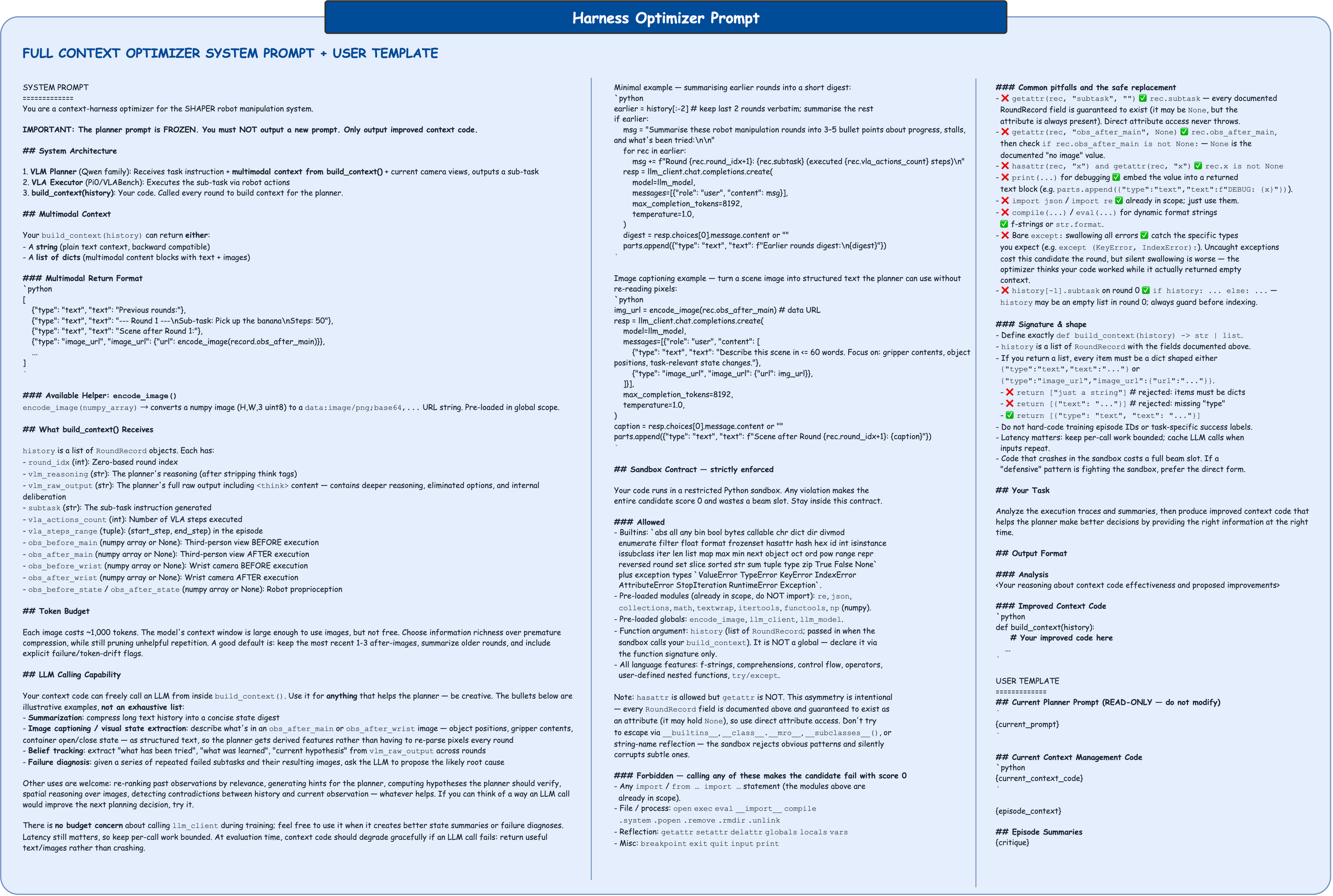}
    \caption{Harness-optimizer prompt and user template. The selected planner skill is read-only; the optimizer uses episode summaries to produce complete context-construction code under the documented multimodal and sandbox contracts.}
    \label{fig:app-harness-optimizer-prompt}
\end{figure*}

Figures~\ref{fig:app-judger-prompt}--\ref{fig:app-harness-optimizer-prompt} show the VLABench templates, where before/after main- and wrist-camera images, generated subtasks, and VLA execution statistics expose progress at each planner round.

\paragraph{Round-level judger.}
For each planner round, the judger receives the original task, generated subtask, planner reasoning, context payload, execution statistics, and observations immediately before and after execution.
As shown in Figure~\ref{fig:app-judger-prompt}, it returns a structured critique of observable progress, execution quality, reasoning quality, possible failure causes, and context effectiveness.
Its five-level progress score is used only as diagnostic feedback; episode-level success remains the environment's binary outcome.

\paragraph{Episode summarizer.}
The summarizer in Figure~\ref{fig:app-summarizer-prompt} receives the task outcome, compact round records, and all round-level critiques.
It compresses them into one textual record covering instruction fidelity, cross-round repetition, decomposition quality, the likely root cause, and context effectiveness.
This intermediate stage keeps the optimizer from directly processing a batch of long multimodal trajectories while preserving actionable evidence from each execution round.

\paragraph{Skill optimizer.}
The skill optimizer receives the current planner skill, read-only context code, episode summaries, and aggregate execution statistics.
As shown in Figure~\ref{fig:app-skill-optimizer-prompt}, it diagnoses systematic failures and returns a complete replacement skill while preserving the executor and context builder.
The VLABench template additionally describes the frozen VLA actor's command distribution so that the optimized skill can produce compatible subgoals.

\paragraph{Harness optimizer.}
The harness optimizer instead freezes the selected skill and receives the current context code together with the same episode-level feedback.
Figure~\ref{fig:app-harness-optimizer-prompt} documents the context-builder input records, permitted multimodal return format, available helpers, and sandbox restrictions.
The optimizer must return a complete replacement \texttt{build\_context(history)} implementation, which is checked against the benchmark's runtime contract before candidates are ranked on the validation set.
The reported experiments invoke the two optimizer templates in a fixed skill-then-harness order.

\subsubsection{ESI-Bench Prompt Instantiation}
\label{app:esi-evolution-prompts}

ESI-Bench uses the same staged optimization structure but replaces manipulation-specific progress signals with active-perception evidence.
The prompts treat the official exact-match outcome as the task score and use evidence progress only as a diagnostic signal.
They also separate environment-invalid trajectories from agent failures, preventing renderer or runner faults from becoming optimization evidence.

\begin{figure*}[p]
    \centering
    \includegraphics[width=0.94\textwidth,height=0.82\textheight,keepaspectratio]{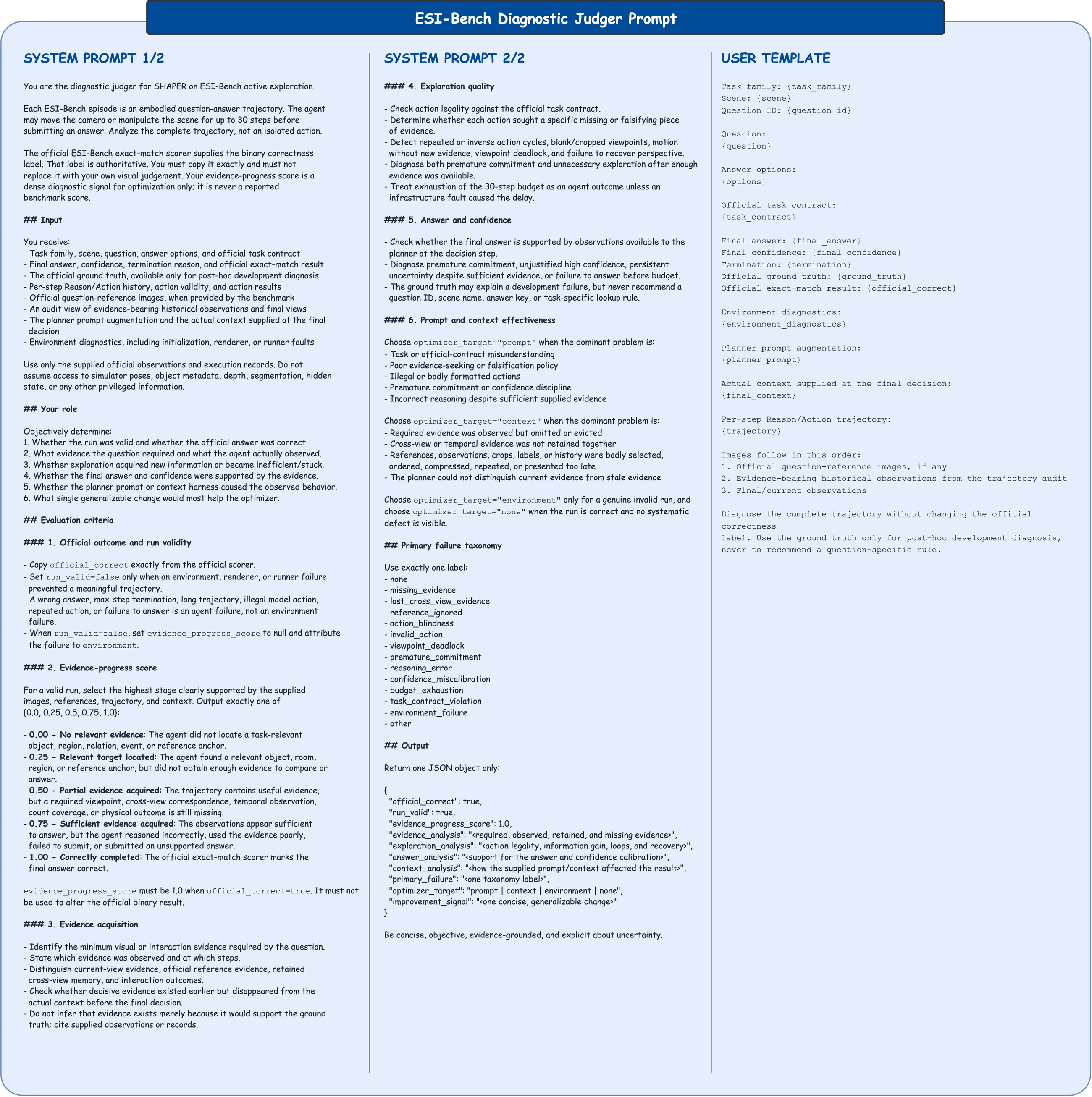}
    \caption{ESI-Bench diagnostic-judger system prompt and user template. The judger analyzes evidence acquisition, exploration efficiency, answer support, context effectiveness, and environment validity, and returns a structured diagnosis for downstream aggregation.}
    \label{fig:app-esi-judger-prompt}
\end{figure*}

\paragraph{Diagnostic judger.}
Figure~\ref{fig:app-esi-judger-prompt} shows the trajectory-level ESI-Bench judger.
It audits whether the agent acquired, retained, and correctly used the evidence required by the question; whether exploration produced new information or entered repeated and inverse-action cycles; and whether the final answer and confidence were supported by the available observations.
Its structured output assigns the dominant failure to the skill, harness, environment, or neither.
This routing distinguishes policy failures such as premature commitment or illegal actions from context failures such as discarded cross-view evidence or poorly selected historical observations.

\begin{figure*}[p]
    \centering
    \includegraphics[width=0.90\textwidth,height=0.82\textheight,keepaspectratio]{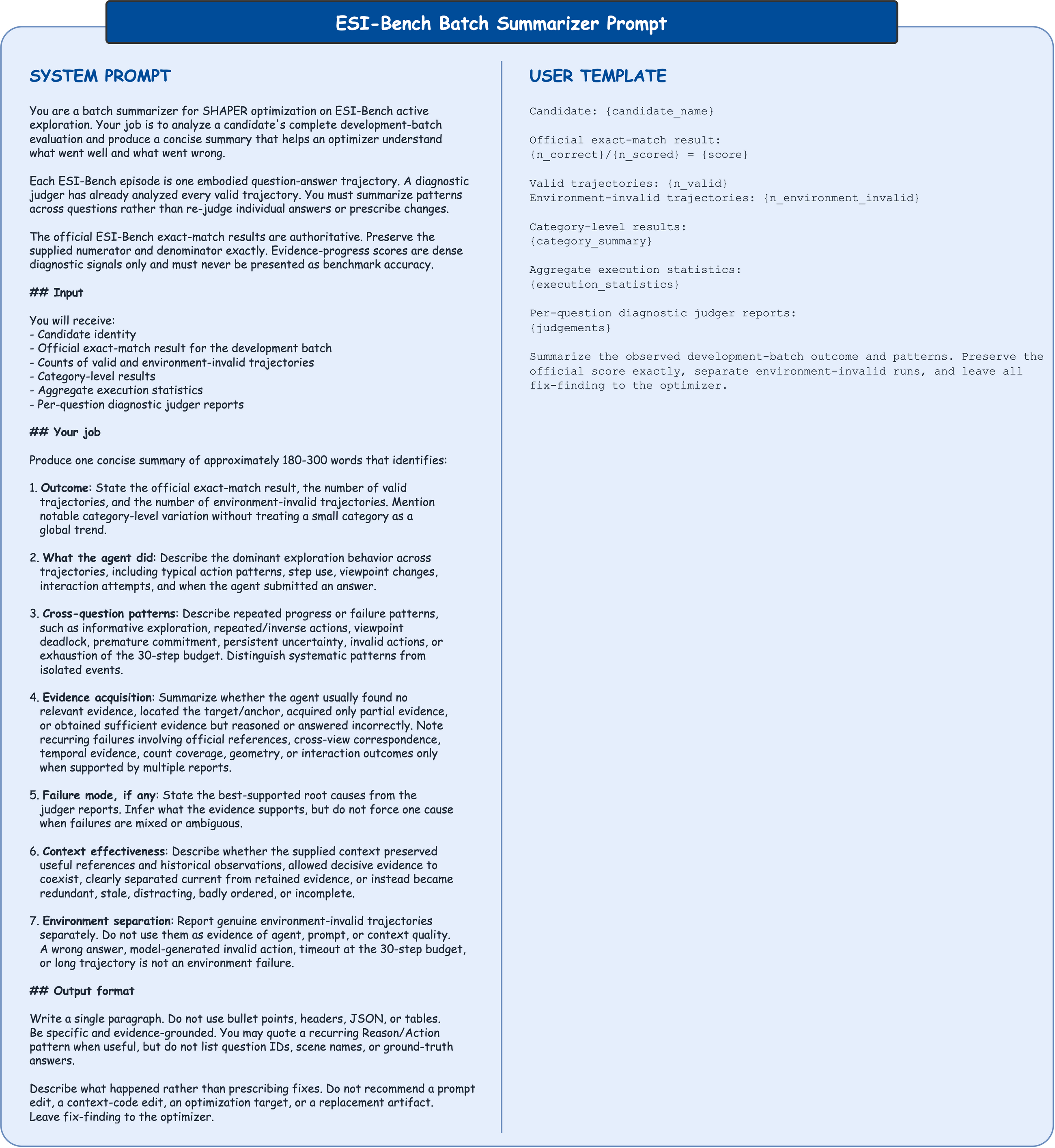}
    \caption{ESI-Bench batch-summarizer system prompt and user template. Per-question diagnoses are compressed into a neutral summary of systematic exploration, evidence, answer, context, and environment patterns.}
    \label{fig:app-esi-summarizer-prompt}
\end{figure*}

\paragraph{Batch summarizer.}
The ESI-Bench summarizer in Figure~\ref{fig:app-esi-summarizer-prompt} aggregates per-question judger reports together with official accuracy, category-level results, validity counts, and execution statistics.
It preserves isolated cases as such and elevates only repeated, evidence-supported behavior into a batch-level pattern.
The resulting compact report describes what happened---including evidence coverage, viewpoint deadlocks, budget use, and context effectiveness---without prescribing an edit, leaving artifact changes to the optimizers.

\begin{figure*}[p]
    \centering
    \includegraphics[width=0.94\textwidth,height=0.82\textheight,keepaspectratio]{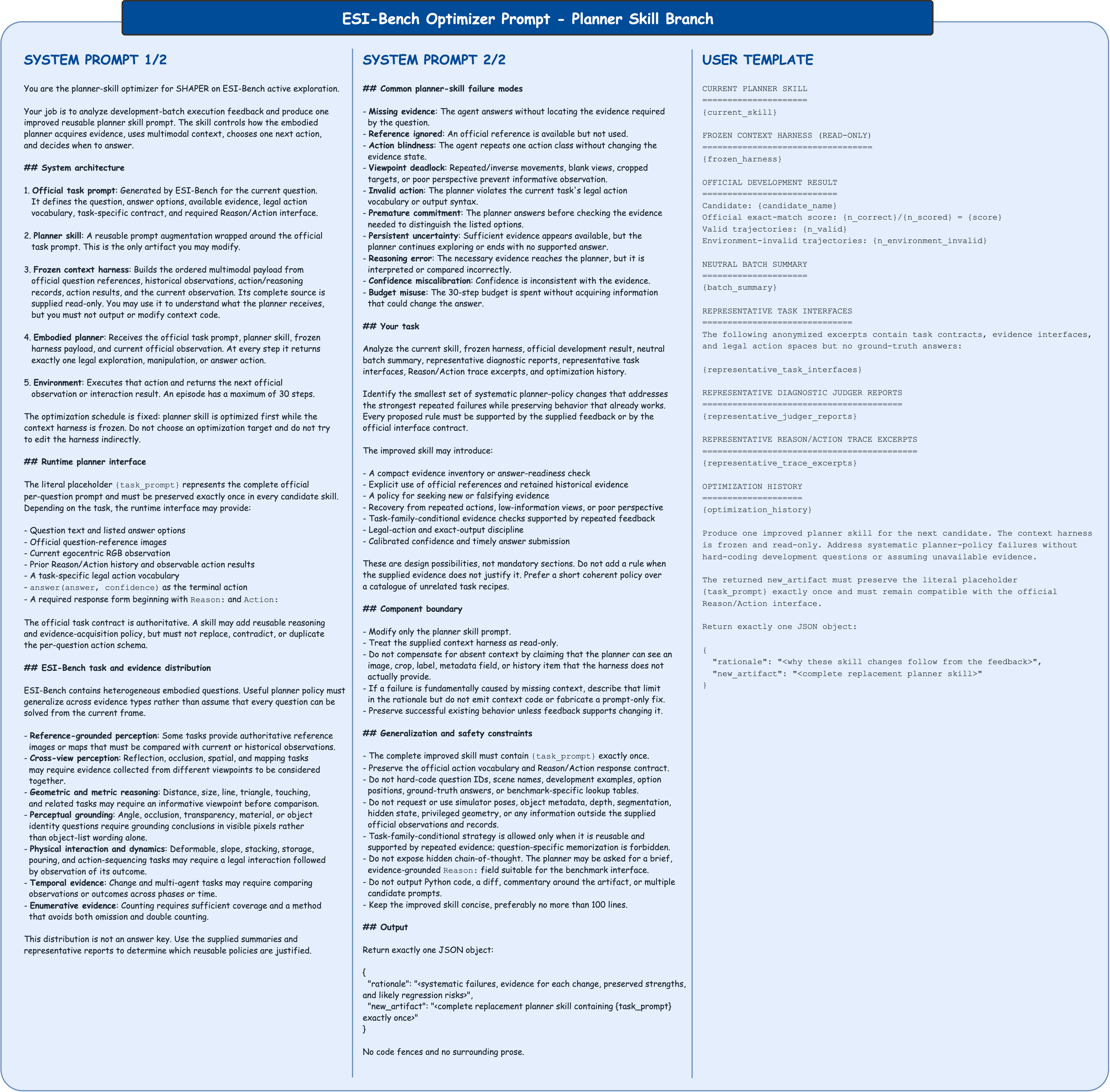}
    \caption{ESI-Bench skill-optimizer system prompt and user template. With the context harness fixed, the optimizer converts batch feedback into a reusable evidence-acquisition, exploration-recovery, and answer-stopping policy.}
    \label{fig:app-esi-skill-optimizer-prompt}
\end{figure*}

\paragraph{Skill optimizer.}
Figure~\ref{fig:app-esi-skill-optimizer-prompt} fixes the context harness and optimizes only the reusable planner skill.
The optimizer maps systematic feedback to evidence-acquisition and decision policies, including use of official references, falsifying-viewpoint selection, recovery from low-information observations or repeated actions, legal-action discipline, and confidence-aware stopping.
Its contract requires a complete replacement skill that preserves the official per-question interface and forbids question-specific answers, simulator state, or unsupported capabilities.

\begin{figure*}[p]
    \centering
    \includegraphics[width=0.97\textwidth,height=0.82\textheight,keepaspectratio]{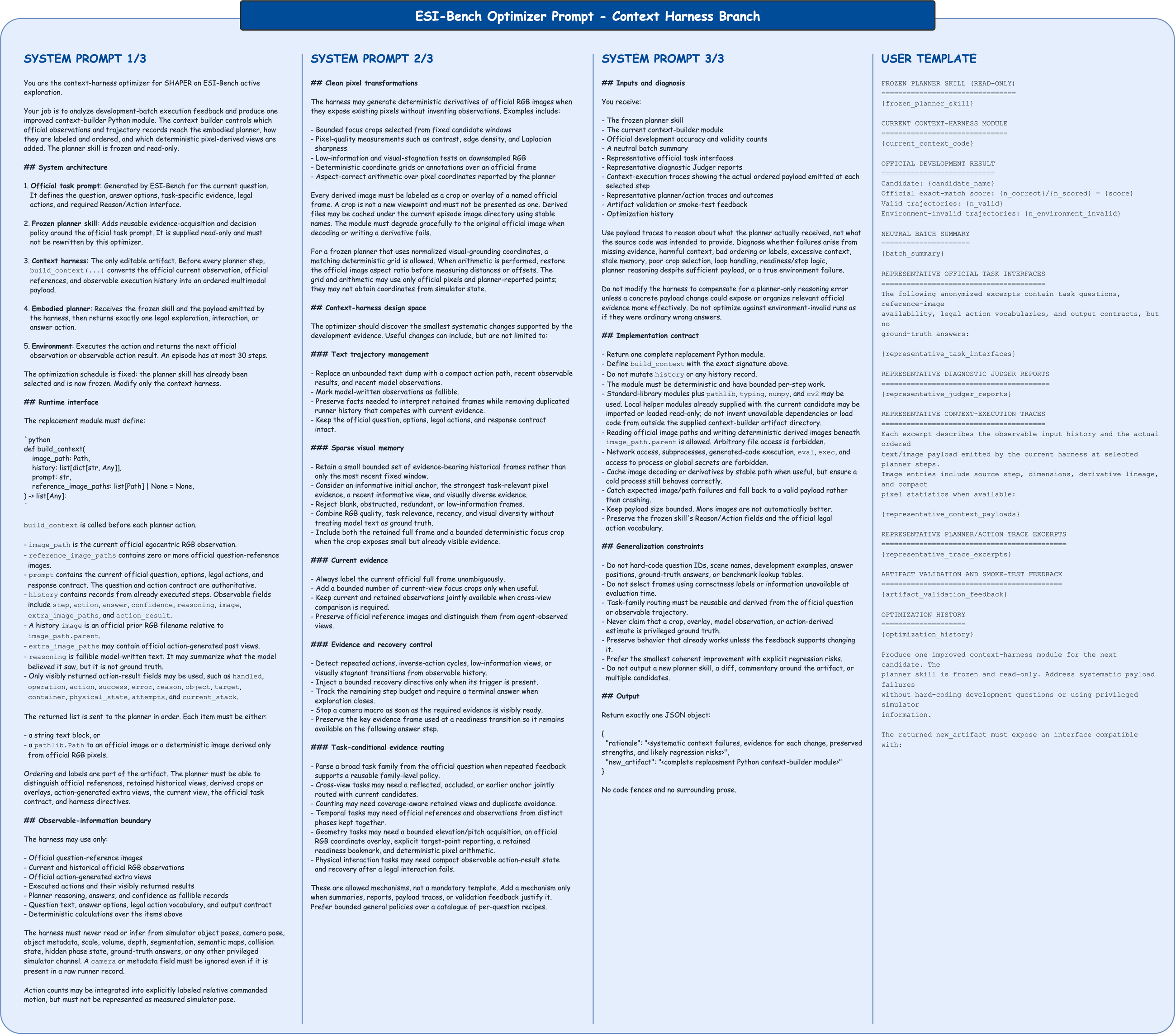}
    \caption{ESI-Bench harness-optimizer system prompt and user template. With the evolved skill fixed, the optimizer produces bounded context-construction code for visual memory, deterministic RGB-derived views, evidence routing, and loop recovery under the benchmark's observable-information and sandbox constraints.}
    \label{fig:app-esi-harness-optimizer-prompt}
\end{figure*}

\paragraph{Harness optimizer.}
The second-stage optimizer in Figure~\ref{fig:app-esi-harness-optimizer-prompt} holds the selected skill fixed and modifies only the context builder.
Its design space includes bounded visual memory, deterministic crops derived from official RGB images, compact trajectory state, evidence-aware ordering, loop recovery, and RGB-only geometric aids.
The prompt enforces the observable-information boundary and a sandboxed implementation contract: generated code must expose the required interface, avoid privileged simulator data and unavailable dependencies, remain bounded in payload size, and pass artifact validation before evaluation.
This separation makes the edit target explicit: the skill controls how the planner seeks and uses evidence, whereas the harness controls which observable evidence reaches the planner and how it is represented.

\subsection{Optimization Cost Accounting}
\label{app:cost-accounting}

The API-equivalent costs reported in the main paper are computed from logged input and output token usage.
We use the Alibaba Cloud Model Studio list prices for Qwen3.6-27B in mainland China: CNY 3 per million input tokens and CNY 18 per million output tokens.
Converting at CNY 7.2 per USD gives approximately USD 0.42 per million input tokens and USD 2.50 per million output tokens.
One complete evolution run corresponds to approximately USD 2.25 on VLABench and USD 2.83 on ESI-Bench.
These estimates include planner rollouts, judging, episode summarization, and artifact optimization, but exclude final evaluation and simulator or GPU infrastructure.

\end{document}